\documentclass[journal]{IEEEtran}
\usepackage{amsmath,amsfonts}
\usepackage{algorithmic}
\usepackage{array}
\usepackage{textcomp}

\usepackage{url}
\usepackage{verbatim}
\usepackage{graphicx}
\def\BibTeX{{\rm B\kern-.05em{\sc i\kern-.025em b}\kern-.08em
    T\kern-.1667em\lower.7ex\hbox{E}\kern-.125emX}}
\usepackage{balance}

\usepackage{utfsym} 
\usepackage{array}
\usepackage{booktabs}
\usepackage{multirow}
\usepackage{threeparttable}
\usepackage{tabularx}
\usepackage{colortbl}
\usepackage[hidelinks]{hyperref}

\definecolor{linkblue}{RGB}{0, 150, 200}
\definecolor{deepviolet}{RGB}{90, 60, 150}
\definecolor{darkgreen}{rgb}{0.0, 0.5, 0.0}
\definecolor{darkred}{rgb}{0.5, 0.0, 0.0}
\definecolor{darkgreen}{rgb}{0.0, 0.5, 0.0}
\definecolor{darkred}{rgb}{0.5, 0.0, 0.0}

\definecolor{softgreen}{RGB}{0,100,0}
\definecolor{softred}{RGB}{150,0,0}
\definecolor{problue}{RGB}{90, 100, 150}
\definecolor{softgrey}{RGB}{128,128,128}

\definecolor{reactive}{RGB}{120,120,120}
\definecolor{active}{RGB}{235,182,45}
\definecolor{proactive}{RGB}{79,15,242}

\definecolor{myblue}{RGB}{233, 241, 249}
\definecolor{mygray}{RGB}{99, 110, 114}
\definecolor{myred}{RGB}{255, 118, 117}
\definecolor{myyellow}{RGB}{255, 234, 167}
\definecolor{mygreen}{RGB}{216, 226, 204}
\definecolor{background_gray}{HTML}{FAFAFA}
\definecolor{frame_gray}{HTML}{424242}
\renewcommand{\thetable}{\arabic{table}} % 保证表格编号是全局连续的

\definecolor{background_blue}{HTML}{E3F2FD}
\definecolor{frame_blue}{HTML}{1565C0}

\definecolor{background_green}{HTML}{E8F5E9}
\definecolor{frame_green}{HTML}{2E7D32}

\definecolor{background_red}{HTML}{FFEBEE}
\definecolor{frame_red}{HTML}{C62828}

\definecolor{highlight_gray}{HTML}{E0E0E0}
\definecolor{highlight_yellow}{HTML}{FFF59D}
\definecolor{elephantgray}{rgb}{0.65,0.65,0.65}

\usepackage{tikz}

\usepackage[utf8]{inputenc} % allow utf-8 input
\usepackage[T1]{fontenc}    % use 8-bit T1 fonts
\usepackage{hyperref}       % hyperlinks
\usepackage{url}            % simple URL typesetting
\usepackage{booktabs}       % professional-quality tables
\usepackage{amsfonts}       % blackboard math symbols
\usepackage{nicefrac}       % compact symbols for 1/2, etc.

\usepackage{xcolor}         % colors
\usepackage[utf8]{inputenc} % allow utf-8 input
\usepackage[T1]{fontenc}    % use 8-bit T1 fonts
\usepackage{hyperref}       % hyperlinks
\usepackage{url}            % simple URL typesetting
\usepackage{booktabs}       % professional-quality tables
\usepackage{amsfonts}       % blackboard math symbols
\usepackage{nicefrac}       % compact symbols for 1/2, etc.
\usepackage{microtype}      % microtypography
\usepackage{xcolor}         % colors
\usepackage[utf8]{inputenc}
\usepackage{color}
\usepackage{graphicx} % 为了使用 \resizebox

\usepackage{booktabs}
\usepackage{multirow}
\usepackage{tabularx}

\usepackage{threeparttable}
\usepackage{tcolorbox}

\usepackage{colortbl}
\usepackage{calligra}     % 花体字支持（需要安装字体
\usepackage{listings}
\usepackage{amsmath}
\usepackage{tabularx}
\usepackage{booktabs}
\usepackage{makecell}  % 可选，用于自动换行

\usepackage{graphicx}
\usepackage{amsmath}
\usepackage{enumitem}
\usepackage{mathrsfs}
\usepackage{float}
\usepackage{multirow} 
\usepackage{booktabs}

\usepackage[table]{xcolor} % 注意 table 选项
\usepackage{colortbl}       % 部分情况下需显式加载

\usepackage{amssymb} % 提供 \blacklozenge 和 \lozenge
\usepackage{dsfont}
\usepackage{threeparttable} % 新增包：支持表格脚注
\definecolor{linkblue}{RGB}{0, 150, 200}
\definecolor{deepviolet}{RGB}{90, 60, 150}

\usepackage{siunitx}
\usepackage{utfsym}  % 提供 \ding 命令
\usepackage{fontawesome}
\usepackage{xcolor}

\definecolor{softgreen}{RGB}{0,100,0}
\definecolor{softred}{RGB}{150,0,0}
\definecolor{problue}{RGB}{90, 100, 150}
\definecolor{softgrey}{RGB}{128,128,128}

\definecolor{reactive}{RGB}{120,120,120}
\definecolor{active}{RGB}{235,182,45}
\definecolor{proactive}{RGB}{79,15,242}

\definecolor{myblue}{RGB}{233, 241, 249}
\definecolor{mygray}{RGB}{99, 110, 114}
\definecolor{myred}{RGB}{255, 118, 117}
\definecolor{myyellow}{RGB}{255, 234, 167}
\definecolor{mygreen}{RGB}{216, 226, 204}

\usepackage{tikz}

\usetikzlibrary{decorations.pathmorphing}
\tikzset{gral/.style={black, line width=0.3mm}}

\newtcolorbox{codebox}[1][]{
  colback=gray!5!white,
  colframe=black,
  boxrule=0.5pt,
  arc=4pt,
  top=3pt,
  bottom=3pt,
  left=6pt,
  right=6pt,
  boxsep=5pt,
  listing only,
  breakable, % This option allows the box to break across pages
  listing options={
    basicstyle=\ttfamily,
    language=Python,
    morekeywords={\quad} 
  },
  #1 % This allows for additional options to be passed at the point of use                                          
}

\definecolor{background_gray}{HTML}{FAFAFA}
\definecolor{frame_gray}{HTML}{424242}
\renewcommand{\thetable}{\arabic{table}} % 保证表格编号是全局连续的

\definecolor{background_blue}{HTML}{E3F2FD}
\definecolor{frame_blue}{HTML}{1565C0}

\definecolor{background_green}{HTML}{E8F5E9}
\definecolor{frame_green}{HTML}{2E7D32}

\definecolor{background_red}{HTML}{FFEBEE}
\definecolor{frame_red}{HTML}{C62828}

\definecolor{highlight_gray}{HTML}{E0E0E0}
\definecolor{highlight_yellow}{HTML}{FFF59D}
\definecolor{elephantgray}{rgb}{0.65,0.65,0.65}

\definecolor{darkgreen}{rgb}{0.0, 0.5, 0.0}
\definecolor{darkred}{rgb}{0.5, 0.0, 0.0}

\usepackage{xcolor}
\usepackage{listings}
\usepackage{mdframed}
\usepackage{bbm}

\makeatletter
\def\@maketitle{%
  \newpage
  \null
  \vskip 1em%
  \begin{center}%
    {\fontsize{24}{26}\selectfont \@title \par}%
    \vskip 1.5em%
    {\normalsize \@author}%
  \end{center}%
  \par
  \vskip 1em%
}
\makeatother

\begin{document}

\title{Transforming Monolithic Foundation Models into Embodied Multi-Agent Architectures for Human-Robot Collaboration}

\author{
\fontsize{12}{14}\selectfont
Nan Sun$^{1}$, Bo Mao$^{2}$,  
Yongchang Li$^{1}$, Chenxu Wang$^{1}$,  
Di Guo$^{2}$ and Huaping Liu$^{1,\dagger}$

\thanks{$^{1}$The author is with the Department of Computer Science and Technology, Tsinghua University, Beijing, 100084, China.}%
\thanks{$^{2}$The author is with the School of Artificial Intelligence,  Beijing University of Posts and Telecommunications, Beijing, 100876, China.}%
\thanks{$^\dagger$Corresponding Authors. {hpliu@tsinghua.edu.cn}}%
%\thanks{$^{*}$Equal Contribution.}%
}

\maketitle

\begin{abstract}
Foundation models have become central to unifying perception and planning in robotics, yet real-world deployment exposes a mismatch between their monolithic assumption that a single model can handle all cognitive functions and the distributed, dynamic nature of practical service workflows. Vision–language models offer strong semantic understanding but lack embodiment-aware action capabilities while relying on hand-crafted skills. Vision–Language–Action policies enable reactive manipulation but remain brittle across embodiments, weak in geometric grounding, and devoid of proactive collaboration mechanisms. These limitations indicate that scaling a single model alone cannot deliver reliable autonomy for service robots operating in human-populated settings. To address this gap, we present InteractGen, an LLM-powered multi-agent framework that decomposes robot intelligence into specialized agents for continuous perception, dependency-aware planning, decision and verification, failure reflection, and dynamic human delegation, treating foundation models as regulated components within a closed-loop collective. Deployed on a heterogeneous robot team and evaluated in a three-month open-use study, InteractGen improves task success, adaptability, and human–robot collaboration, providing evidence that multi-agent orchestration offers a more feasible path toward socially grounded service autonomy than further scaling standalone models.
\end{abstract}


\begin{figure*}[t]
    \centering
    \includegraphics[width=1\linewidth]{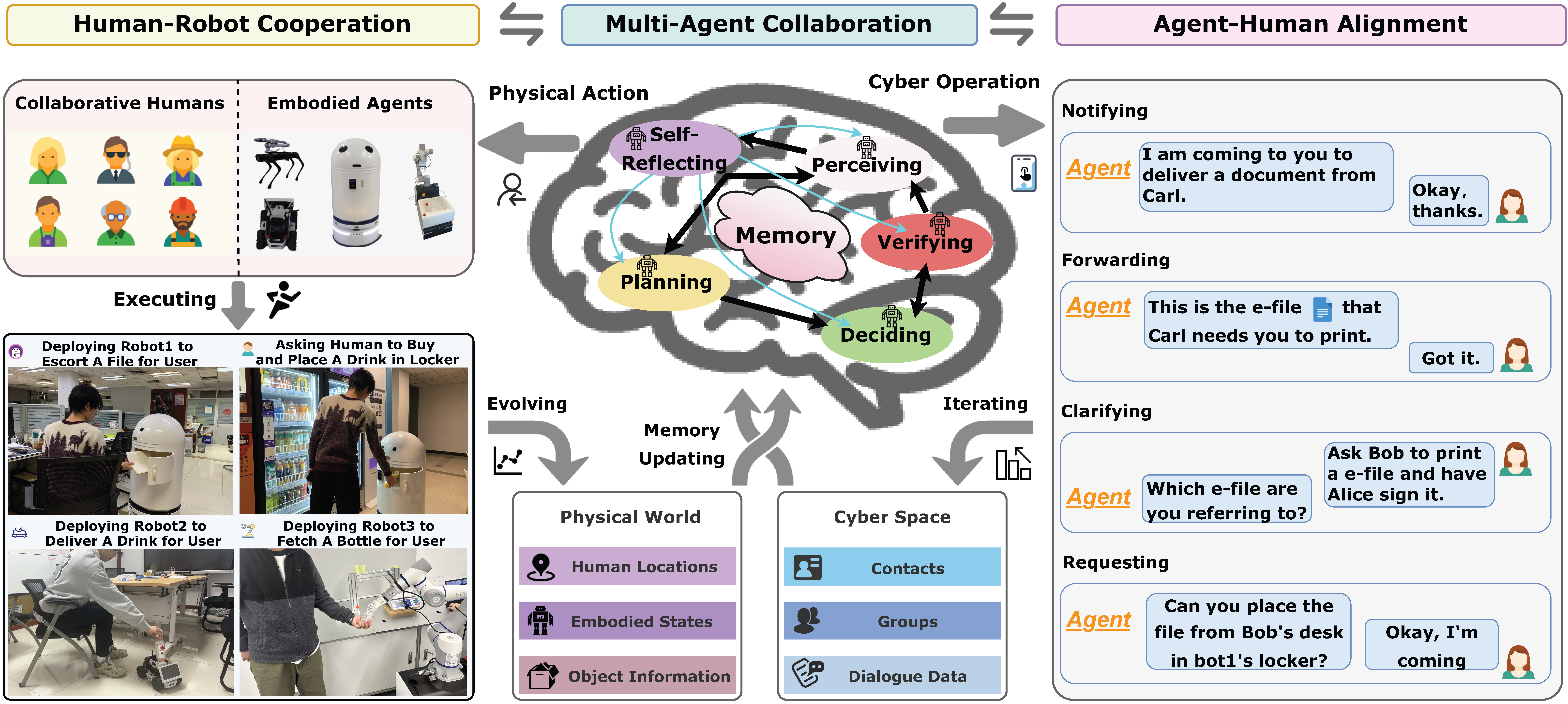}
    \vspace{-3mm}
    \caption{\textbf{A holistic demonstration of how InteractGen coordinates robots and humans in real time.} The framework monitors online and offline environments, performs collaborative reasoning across specialized agents, plans composite workflows, and triggers physical execution through heterogeneous robots. Humans are treated as deployable agents—InteractGen clarifies, notifies, and delegates subtasks when appropriate—enabling socially grounded service autonomy.}

    \label{fig:introduction}
    \vspace{-3mm}
\end{figure*}

\section{Introduction}
\label{sec:introduction}

\IEEEPARstart{F}{oundation} models have rapidly entered the robotics community as a unifying abstraction for perception, language, and action~\cite{zhang2024llmgrop,xu2024setitup,lee2024primethesearch,firoozi2025foundation}, yet their real-world impact on service robots remains far from the autonomy required in human-populated environments. Current robotics foundation models fall into two powerful but fundamentally incomplete families. The first comprises \emph{embodied Vision-Language-Models (VLM)}—such as VeBrain~\cite{luo2025visualembodiedbrainlet}, RoboBrain~\cite{ji2025robobrainunifiedbrainmodel} and MiMo-Embodied~\cite{hao2025mimoembodiedxembodiedfoundationmodel}—which unify multimodal understanding and spatial reasoning, and often demonstrate strong performance on planning. However, these models typically depend on post-training with benchmark-specific adaptations, in which abilities distilled from larger VLMs are operationalized by generating training data using diverse task templates; however, their action components are generally delegated to hand-designed low-level skill libraries, as in SayCan~\cite{ahn2022saycan}, and concrete demonstrations of multi-robot execution, embodiment-aware action, or human-centered deployment remain limited. 

The second family comprises \emph{Vision-Language-Action (VLA) policies}, which primarily fine-tune relatively smaller VLMs on robot action datasets using autoregressive language heads~\cite{Brohan2023RT2VM,Kim2024OpenVLAAO} or diffusion/flow-based action experts~\cite{black2024pi0visionlanguageactionflowmodel,li2024cogactfoundationalvisionlanguageactionmodel,intelligence2025pi05visionlanguageactionmodelopenworld}, achieving promising reactive manipulation across multiple embodiments and scenes. However, deploying a model in \emph{real-world service settings} requires \emph{comprehensive autonomy}: an assistant operating in such environments must simultaneously parse instructions, generate appropriate motion plans, manage uncertainty, and interact fluidly with surrounding human users. Under these demands, current VLAs serve as excellent motor primitives but remain inadequate as assistive systems: they lack persistent task memory, contextual reflection, collaboration awareness, and mechanisms for interacting with humans within realistic workflows. Together, these trends highlight a structural tension: despite substantial isolated capabilities, neither family of foundation models yet realizes the intentional, contextually grounded autonomy required for service robots to deliver meaningful social utility.

These considerations point toward a broader insight: the difficulty of deploying current foundation models in real-world service robots does not arise merely from limited scale or training data, but from a deeper mismatch between their monolithic architectures and the distributed, interdependent nature of actual service tasks. Real-world environments inherently involve multiple humans, robots, tools, and information channels, and therefore require cognitive structures capable of arbitrating uncertainty and coordinating heterogeneous embodiments over long horizons. Evidence from digital multi-agent LLM systems such as AutoGen~\cite{wu2023autogenenablingnextgenllm}, MetaGPT~\cite{Hong2023MetaGPTMP}, and AgentVerse~\cite{chen2023agentversefacilitatingmultiagentcollaboration} reinforces this view: even without embodiment, reliable long-horizon autonomy typically emerges from role decomposition, reflective loops, tool invocation, and explicit verification steps, rather than from a single large model. When physical robots enter the loop—with real-world dynamics, partial observability, safety constraints, and human unpredictability—the limitations of monolithic foundation models become even more pronounced. For the foreseeable future, it is unlikely that any single VLM or VLA claiming to be an embodied foundation model will simultaneously deliver spatial grounding, cross-embodiment skill composition, multi-agent task allocation, safety arbitration, and socially appropriate interaction. What is needed instead is an architectural shift—one that treats foundation models not as monolithic controllers but as powerful \emph{components} embedded within structured multi-agent control systems that support \emph{collaborative embodied reasoning}. Such a design enables the flexibility, interpretability, and competence required for a genuinely general-purpose robotic brain.

Motivated by this perspective, we introduce InteractGen, an LLM-powered multi-agent framework enabling collaborative embodied reasoning through specialized agents built on diverse foundation models (Fig.~\ref{fig:introduction}). InteractGen coordinates heterogeneous robots and humans via five agents for perception, planning, decision making, validation, and reflection. This decomposition extends beyond single-loop paradigms such as ReAct~\cite{yao2023react} and Reflexion~\cite{shinn2023reflexionlanguageagentsverbal}, while avoiding decentralized designs requiring a separate LLM per robot as in EMOS~\cite{chen2025emosembodimentawareheterogeneousmultirobot} which is hard to scale. InteractGen can also invoke API-level tool calls for GUI operations while serving as a unified “robot brain’’ for low-level skill libraries or end-to-end VLA policies. By centralizing high-level reasoning and delegating physical execution to generalist controllers or modular primitives, InteractGen treats foundation models as collaborative components rather than monolithic controllers.

Crucially, InteractGen is designed as a human-centered interactive system that takes humans as \emph{deployable agents} capable of taking on delegated subtasks whenever robots encounter safety, authority, or dexterity limits. This allows the system to recognize its own bounds and proactively solicit human assistance when tasks exceed robotic capability in dynamic environments, pushing embodied applications beyond isolated navigation or manipulation toward realistic service workflows that require joint digital–physical coordination. Across controlled evaluations and a three-month open-use deployment, we observed that even simple mobile robots became markedly more useful when embedded in this human-aware loop—acting not merely as executors but as assistants that could anticipate needs, delegate tasks when appropriate, and engage in extended multi-user workflows. These results suggest that multi-agent orchestration, combined with meaningful human involvement, offers a more promising path to socially grounded autonomy than further scaling monolithic foundation models, underscoring that effective human–robot collaboration is central to the future of service robotics.

% 放在导言区或表格之前

\newcommand{\GreenDot}{%
  \tikz[baseline=-0.6ex]\fill[darkgreen] (0,0) circle (0.8ex);%
}
\newcommand{\GrayBox}{%
  \tikz[baseline=-0.6ex]\draw[fill=gray!20] (0,0) rectangle (0.2,0.2);%
}

\begin{table*}[!t]
\centering
\caption{\textbf{Comparison between InteractGen and prevailing agentic systems}. \textit{Integrated} denotes single-step action generation, whereas \textit{Hierarchical} denotes explicit separation between planning and decision making. \textit{Validation} refers to pre-execution self-checking, and \textit{Reflection} refers to post-execution reasoning. In the \textbf{Interaction Skills} columns, grey rectangles indicate that the system does not explicitly support this capability in its design or operation.}
\label{tab:methodcomparison}
\begin{tabular}{lccccccc}
\toprule
\multirow{2}{*}{Methods} 
  & \multicolumn{4}{c}{\textbf{\textsc{Reasoning Composition}}}
  & \multicolumn{3}{c}{\textbf{\textsc{Interaction Skills}}} \\
\cmidrule(lr){2-5} \cmidrule(l){6-8}
  & Perception 
  & Planning \& Decision 
  & Validation 
  & Reflection 
  & Robot Dispatch 
  & GUI Operation 
  & Human Collaboration\\ 
\midrule

ReAct~\cite{yao2023react} 
  & \textcolor{darkgreen}{\checkmark} 
  & \textcolor{elephantgray}{\textit{Integrated}}
  & \textcolor{darkred}{\usym{2613}} 
  & \textcolor{darkred}{\usym{2613}} 
  & \GreenDot
  & \GrayBox
  & \GrayBox \\

Reflexion~\cite{shinn2023reflexionlanguageagentsverbal} 
  & \textcolor{darkgreen}{\checkmark} 
  & \textcolor{elephantgray}{\textit{Integrated}}
  & \textcolor{darkred}{\usym{2613}} 
  & \textcolor{darkgreen}{\checkmark} 
  & \GreenDot
  & \GrayBox
  & \GrayBox \\

Mobile-Agent-v2~\cite{wang2024mobileagentv2mobiledeviceoperation} 
  & \textcolor{darkgreen}{\checkmark} 
  & \textit{Hierarchical}
  & \textcolor{darkred}{\usym{2613}} 
  & \textcolor{darkgreen}{\checkmark} 
  & \GrayBox
  & \GreenDot
  & \GrayBox \\

KnowNo~\cite{ren2023robotsaskhelpuncertainty} 
  & \textcolor{darkgreen}{\checkmark} 
  & \textcolor{elephantgray}{\textit{Integrated}}
  & \textcolor{darkgreen}{\checkmark} 
  & \textcolor{darkred}{\usym{2613}} 
  & \GreenDot
  & \GrayBox
  & \textit{Clarification} \\

SMART-LLM~\cite{kannan2024smart} 
  & \textcolor{darkgreen}{\checkmark} 
  & \textcolor{elephantgray}{\textit{Integrated}}
  & \textcolor{darkred}{\usym{2613}} 
  & \textcolor{darkred}{\usym{2613}} 
  & \GreenDot
  & \GrayBox
  & \GrayBox \\

CoELA~\cite{zhang2024buildingcooperativeembodiedagents} 
  & \textcolor{darkgreen}{\checkmark} 
  & \textit{Hierarchical}
  & \textcolor{darkgreen}{\checkmark} 
  & \textcolor{darkred}{\usym{2613}} 
  & \GreenDot
  & \GrayBox
  & \GrayBox \\

CaPo~\cite{liu2025capocooperativeplanoptimization} 
  & \textcolor{darkgreen}{\checkmark} 
  & \textit{Hierarchical}
  & \textcolor{darkgreen}{\checkmark} 
  & \textcolor{darkred}{\usym{2613}} 
  & \GreenDot
  & \GrayBox
  & \GrayBox \\

RoCo~\cite{mandi2023rocodialecticmultirobotcollaboration} 
  & \textcolor{darkgreen}{\checkmark} 
  & \textcolor{elephantgray}{\textit{Integrated}}
  & \textcolor{darkred}{\usym{2613}} 
  & \textcolor{darkred}{\usym{2613}} 
  & \GreenDot
  & \GrayBox
  & \textit{Supervision} \\

Lip-LLM~\cite{obata2024lipllmintegratinglinearprogramming} 
  & \textcolor{darkgreen}{\checkmark} 
  & \textit{Hierarchical}
  & \textcolor{darkgreen}{\checkmark} 
  & \textcolor{darkred}{\usym{2613}} 
  & \GreenDot
  & \GrayBox
  & \GrayBox \\

EMOS~\cite{chen2025emosembodimentawareheterogeneousmultirobot} 
  & \textcolor{darkgreen}{\checkmark} 
  & \textcolor{elephantgray}{\textit{Integrated}}
  & \textcolor{darkgreen}{\checkmark} 
  & \textcolor{darkred}{\usym{2613}} 
  & \GreenDot
  & \GrayBox
  & \GrayBox \\

LaMMA-P~\cite{zhang2025lammapgeneralizablemultiagentlonghorizon} 
  & \textcolor{darkgreen}{\checkmark} 
  & \textit{Hierarchical}
  & \textcolor{darkgreen}{\checkmark} 
  & \textcolor{darkred}{\usym{2613}} 
  & \GreenDot
  & \GrayBox
  & \GrayBox \\

HMCF~\cite{li2025hmcfhumanintheloopmultirobotcollaboration} 
  & \textcolor{darkgreen}{\checkmark} 
  & \textit{Hierarchical}
  & \textcolor{darkgreen}{\checkmark} 
  & \textcolor{darkgreen}{\checkmark} 
  & \GreenDot
  & \GrayBox
  & \textit{Supervision} \\

\midrule
InteractGen (Ours) 
  & \textcolor{darkgreen}{\checkmark} 
  & \textit{Hierarchical}
  & \textcolor{darkgreen}{\checkmark} 
  & \textcolor{darkgreen}{\checkmark} 
  & \GreenDot
  & \GreenDot
  & \textit{Subtask Delegation} \\

\bottomrule
\end{tabular}
\end{table*}

In summary, this work makes four key contributions: 
\begin{itemize} 
\item \textbf{Insights into designing multi-agent frameworks built on foundation models.} We identify structural weaknesses in single-model robotic foundations and demonstrate how different classes of foundation models can be systematically embedded into multi-agent architectures tailored for embodied tasks, providing design principles for scalable, interpretable, and robust embodied systems. 

\item \textbf{An LLM-driven multi-agent architecture for real-world multi-user tasks.} We introduce \textbf{InteractGen}, a framework that decomposes robot intelligence into specialized agents for perception, planning, decision making, verification, and reflection—treating foundation models as modular components rather than monolithic controllers to enable collaborative embodied reasoning. 

\item \textbf{A novel human-as-deployable-agent paradigm for effective Human-Robot Collaboration.} We formalize a paradigm in which humans serve as \emph{deployable agents} who can be delegated subtasks when robots face safety, authority, or dexterity limits, enabling robust teaming in realistic multi-user environments with dynamic human participation and hybrid digital--physical workflows. 

\item \textbf{Large-scale real-world deployment demonstrating generalization and social utility.} We deploy InteractGen across a heterogeneous robot team and conduct extensive experiments plus a three-month open-use study with non-expert occupants, demonstrating strong cross-embodiment generalization, reliable human-robot collaboration, and substantial social value in real-world service environments. \end{itemize}

\section{Related Work}
\label{sec:related}

\subsection{Foundation Models for Robotics}
Foundation Models for Robotics are typically seen as large multimodal backbones supporting perception, language understanding, and action. Existing approaches largely follow two trajectories. VLMs such as CLIP~\cite{radford2021learningtransferablevisualmodels} and PaLI-Gemma~\cite{beyer2024paligemmaversatile3bvlm} provide strong scene understanding and semantic grounding. Embodied variants including VeBrain~\cite{luo2025visualembodiedbrainlet}, RoboBrain~\cite{ji2025robobrainunifiedbrainmodel}, and MiMo-Embodied~\cite{hao2025mimoembodiedxembodiedfoundationmodel} extend these capabilities into physical environments, often through distillation and skill-conditioned prompting, as in SayCan~\cite{ahn2022saycan}. That being said, existing VLMs largely rely on hand-crafted controllers and exhibit limited robustness in action grounding.

VLA policies such as RT-2~\cite{Brohan2023RT2VM}, $\pi_0$~\cite{black2024pi0visionlanguageactionflowmodel}, OpenVLA~\cite{Kim2024OpenVLAAO}, and CogAct~\cite{li2024cogactfoundationalvisionlanguageactionmodel} augment VLMs with end-to-end action heads. Despite improved perception–action coupling, these policies remain reactive, with limited task memory, weak uncertainty modeling, and no mechanisms for human involvement or multi-agent coordination. Even in structured VLN and VLA benchmarks, both VLMs and VLAs remain fundamentally single-agent systems.

In summary, current foundation models fall short of the distributed, interactive cognition required in multi-user, real-world service environments, motivating their integration as modular components within more comprehensive frameworks.

% ------------------------------------------------------

\subsection{LLM-Powered Multi-Agent Frameworks}
Recent work in virtual domains increasingly adopts LLM multi-agent architectures, where specialized agents coordinate via explicit role decomposition. Frameworks such as AutoGen~\cite{wu2023autogenenablingnextgenllm}, MetaGPT~\cite{Hong2023MetaGPTMP}, and AgentVerse~\cite{chen2023agentversefacilitatingmultiagentcollaboration} show advantages in long-horizon reasoning and verification over single-agent methods such as ReAct~\cite{yao2023react} and Reflexion~\cite{shinn2023reflexionlanguageagentsverbal}. Yet these systems remain fundamentally disembodied, without grounding in perception, physical constraints, or actuation.

While such virtual-domain systems highlight the importance of structured role decomposition, their embodied counterparts remain far more~\cite{1668250,10183654}. Existing attempts toward embodied multi-agent systems make only partial progress.
 SMART-LLM~\cite{kannan2024smart} integrates multiple modules but lacks explicit role structure; CoELA~\cite{zhang2024buildingcooperativeembodiedagents} and CaPo~\cite{liu2025capocooperativeplanoptimization} support multi-robot collaboration but overlook human interaction and hierarchical verification; LaMMA-P~\cite{zhang2025lammapgeneralizablemultiagentlonghorizon} focuses on simulated multi-agent planning without real-world actuation.

In contrast, InteractGen unifies specialized foundation models across five reasoning roles and couples them with physical execution pipelines and human-facing interfaces. This enables embodied multi-agent coordination that dispatches robots, uses GUI tools, verifies execution, and dynamically delegates subtasks to humans, offering broader reasoning coverage and richer interaction skills than prior work (see Table~\ref{tab:methodcomparison}) .

% ------------------------------------------------------
\begin{figure*}[!htbp]
    \centering
    \includegraphics[width=1\linewidth]{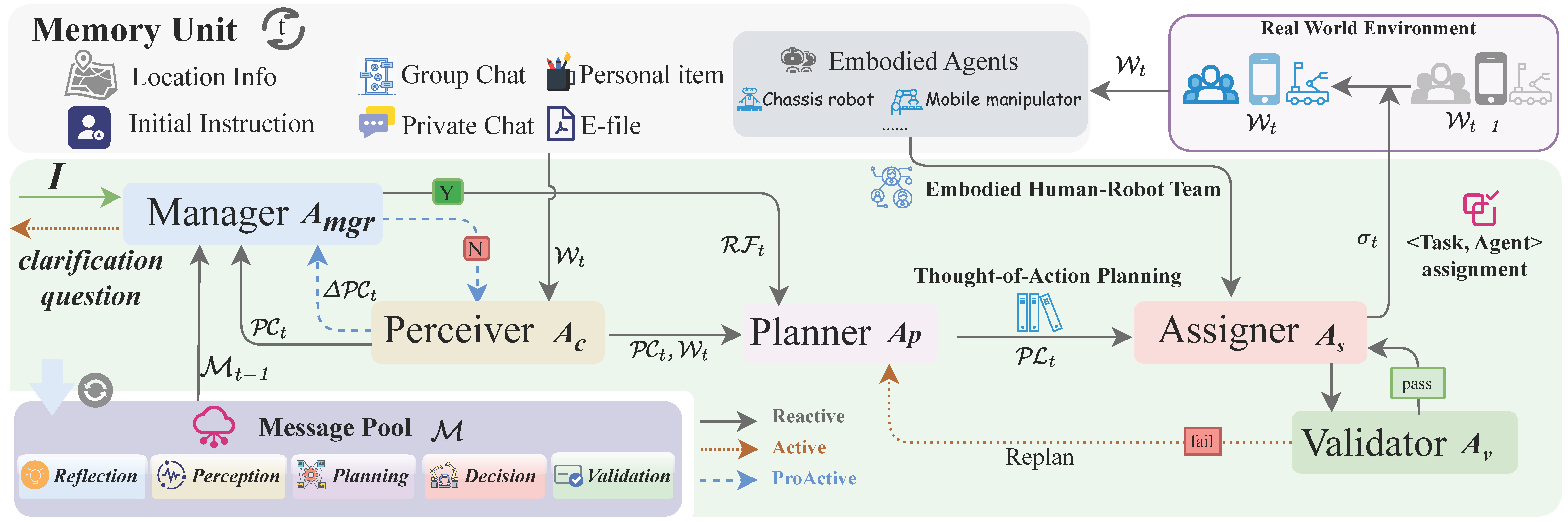}
    \vspace{-3mm}
    \caption{\textbf{Overview of the InteractGen architecture}. InteractGen naturally supports three operating modes that emerge to handle long-horizon, human-centered scenarios. 
    \textbf{\textsc{\textcolor{reactive}{Reactive}:}} Perceiver perceives task-relevant information and signals Planner to generates a Thought-of-Action plan; Assigner assigns actions to suitable agents. 
     \textbf{\textsc{\textcolor{active}{active}:}} Manager triggers clarification for ambiguous cases and Validator's validation elicits re-planning, avoiding rigid generate-then-execute patterns. 
     \textbf{\textsc{\textcolor{problue}{Proactive}:}} Manager reflects, corrects prior reasoning, and reactivates Perceiver for proactive reasoning. These modes enable interactive coordination with humans and robots in dynamic environments. The \textit{act–fail–reflect–replan} mechanism in Proactive Mode greatly enhances human–robot collaboration.}
    \label{fig:agent}
\end{figure*}

\subsection{Human--Robot Collaboration in Agentic Systems}
Human–robot collaboration in agentic systems typically follows two paradigms. The human-as-clarifier model involves humans resolving ambiguities during execution~\cite{ren2023robotsaskhelpuncertainty}. Although effective for disambiguation, this restricts humans to simple query answering and becomes inefficient in long-horizon embodied tasks~\cite{banerjee2025askaskhumanintheloopcontextual}. The human-as-supervisor paradigm places humans in high-level oversight roles, but continuous monitoring imposes substantial cognitive load~\cite{mandi2023rocodialecticmultirobotcollaboration}.

A third emerging paradigm treats humans as deployable agents for subtasks beyond robotic capability. Early systems such as CoBots~\cite{veloso2018symbotic} and modern assistants like AssistantX~\cite{sun2025assistantxllmpoweredproactiveassistant} demonstrate proactive human involvement, but existing implementations remain narrow, rule-based, or domain-specific. InteractGen advances this paradigm by explicitly integrating humans into the multi-agent loop as active collaborators: robots can delegate subtasks and request human actions for tasks they cannot safely, legally, or dexterously perform, thereby achieving better overall task outcomes. This enables more natural, scalable, and socially grounded teamwork in dynamic, multi-user service workflows where robot-only autonomy is insufficient.

\section{Methodology of InteractGen}
\label{sec:method}

\subsection{Agent Specifications of InteractGen}
\label{sec:agent}

InteractGen consists of five LLM-driven agents operating over a shared message
pool and a persistent world-state memory (see Fig.~\ref{fig:agent}). Details of the choices on foundation models for each agent, together with the corresponding insights, are further presented in Section \ref{sec:experiment} and Section \ref{sec:discussion}.

\begin{figure*}[!htbp]
    \centering
    \includegraphics[width=1\linewidth]{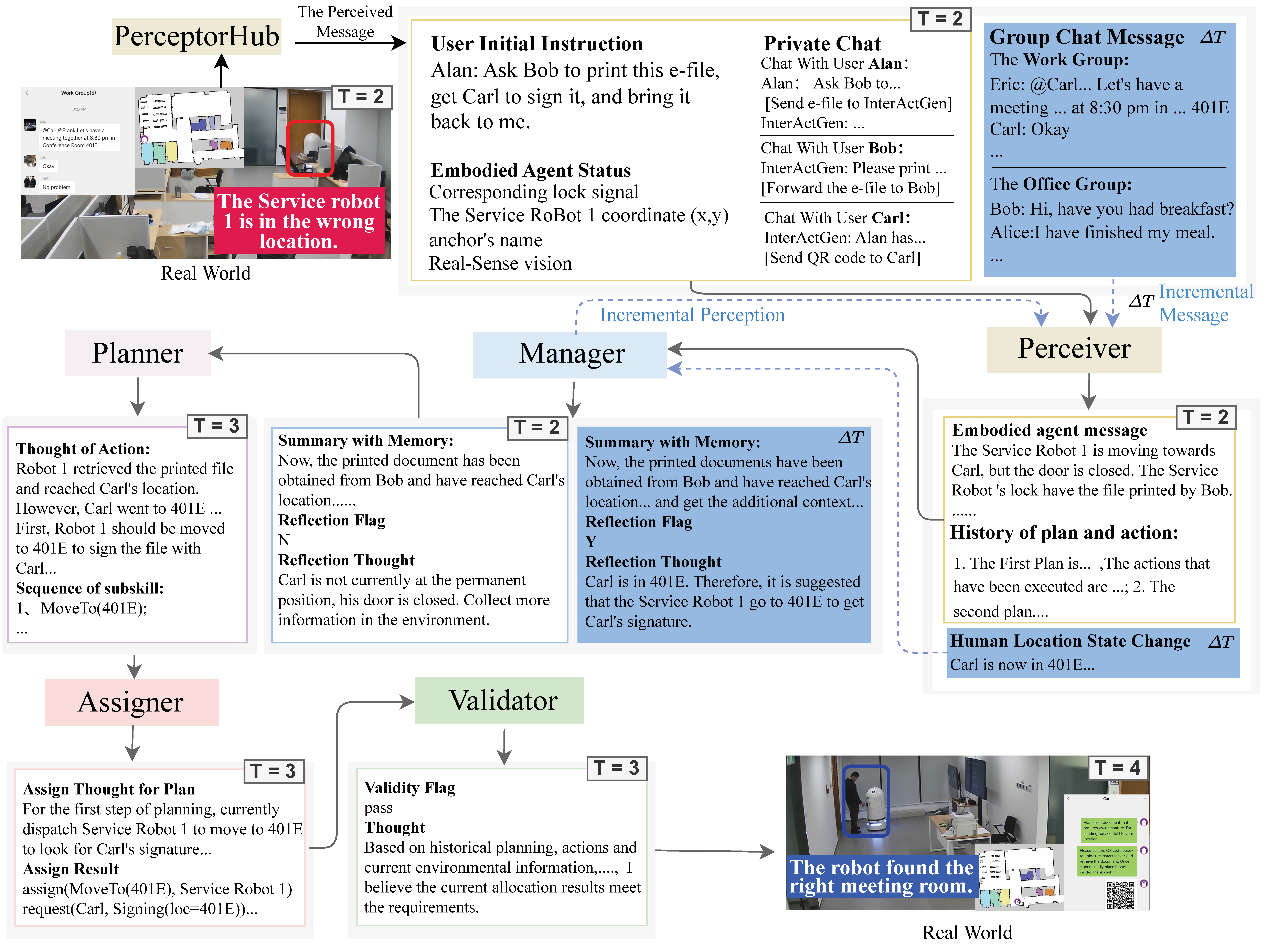}
    \vspace{-3mm}
\caption{
\textbf{A proactive case of InteractGen reasoning and coordination.} 
The system executes the instruction through a modular pipeline. The PerceptorHub collects raw sensor and chat signals, which are incrementally updated by the Perceiver. The Planner generates dependency-aware Thought-of-Action steps, while the Manager monitors progress and triggers reflection when inconsistencies arise (e.g., Carl not at his usual location). The Assigner distributes subtasks to robots or humans, and the Validator checks feasibility before execution. \textbf{This act–reflect–replan loop enables InteractGen to adapt to dynamic human availability and environmental changes, ensuring reliable multi-user task execution in real-world settings.}
}
    \label{fig:proactive}
    \vspace{-4mm}
\end{figure*}

\begin{figure*}[!htbp]
    \centering
    \includegraphics[width=1.0\linewidth]{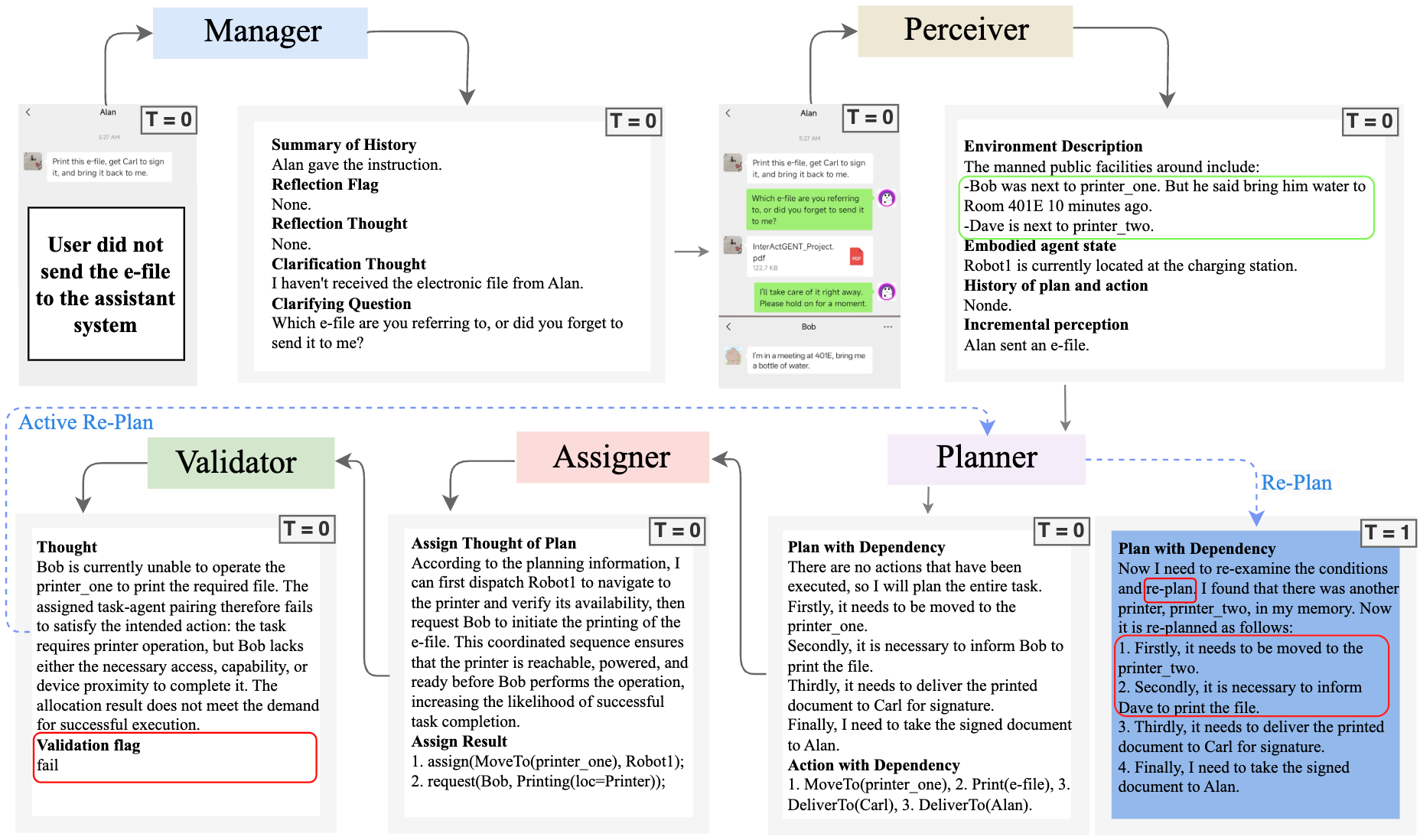}
    \vspace{-3mm}
    \caption{\textbf{Example of active reasoning in InteractGen}. The Manager raises a clarification question when input is ambiguous. After validation fails, the Planner re-generates a feasible plan using updated context. This showcases the system’s ability to actively adapt before execution errors occur.}
    \label{fig:activereasoning}
    \vspace{-3mm}
\end{figure*}

%--------------------------------------
\subsubsection{Agent Manager ($A_{\mathrm{mgr}}$) -- Reflection \& Clarification}

\(A_{\mathrm{mgr}}\) is the reflection agent, whose output \(\mathcal{RF}_{t}\) consists of three components:  
(i) a binary success flag \texttt{\textcolor{softgreen}{Y}/\textcolor{softred}{N}} indicating whether the previous plan \(\mathcal{PL}_{t-1}\) and execution \(\sigma_{t-1}\) produced the desired outcome given perception \(\mathcal{PC}_{t}\);  
(ii) a reflection thought explaining the result and providing corrective guidance; and  
(iii) a concise summary of \(\mathcal{M}_{t-1}\).  
If the flag is \texttt{\textcolor{softgreen}{Y}}, \(A_{\mathrm{mgr}}\) signals \(A_{p}\) to proceed with planning; if \texttt{\textcolor{softred}{N}}, it activates \(A_{c}\) to gather additional context \textcolor{problue}{\(\Delta \mathcal{PC}_{t}\)}, enabling proactive re-planning. As illustrated in Fig.~\ref{fig:proactive}, the Manager’s self-reflection and flexible communication allow the system to adapt to real-world disturbances and autonomously revise the plan for robust execution.
Its process can be formalized as:
\begin{equation}
(\mathcal{I},\, \mathcal{PL}_{t-1},\, \sigma_{t-1},\, \mathcal{M}_{t-1},\, \mathcal{PC}_{t})
\xrightarrow{\text{\tiny $A_{\mathrm \tiny{mgr}}$}}
\mathcal{RF}_{t}
\end{equation}

\(A_{\mathrm{mgr}}\) is also equipped with a \textit{clarification module}, invoked upon receiving the initial user instruction, that returns either \{\texttt{clear}\} or \{\texttt{unclear + clarification question}\}; if the result is \texttt{unclear}, the system relays the question directly to the user for active clarification, as shown in Fig.~\ref{fig:activereasoning}.

%--------------------------------------
\subsubsection{Agent Perceiver ($A_{c}$) -- World-State Perception} 

\(A_{c}\) receives prior world state \(\mathcal{W}_{t-1}\), plan \(\mathcal{PL}_{t-1}\), and executed actions \(\sigma_{t-1}\) (including completion signals from robots), and produces perception output \(\mathcal{PC}_{t}\) while updating \(\mathcal{W}_{t-1}\) to \(\mathcal{W}_{t}\). \(\mathcal{PC}_{t}\) encodes structured representations of task-relevant entities and environmental conditions for iteration \(t\). If \(A_{\mathrm{mgr}}\) reports failure, \(A_{c}\) is prompted to invoke perceptors from the integrated \emph{PerceptorHub}—which unifies modalities such as LiDAR-based localization, RealSense vision, and online dialogue or GUI APIs—to produce reflection-driven, real-time proactive observations \textcolor{problue}{\(\Delta \mathcal{PC}_{t}\)} for re-planning. 
The operation of this agent can be formalized as:
\begin{equation}
(\mathcal{I},\, \mathcal{W}_{t-1},\, \mathcal{PL}_{t-1},\, \sigma_{t-1})
\xrightarrow{\text{\tiny $A_{p}$}}
\mathcal{PC}_{t}
\end{equation}

%--------------------------------------
\subsubsection{Agent Planner ($A_{p}$) -- Thought-of-Action Planning}

Given the current perception \(\mathcal{PC}_{t}\) and reflection \(\mathcal{RF}_{t}\), the planner \(A_{p}\) produces a \emph{Thought-of-Action} plan \(\mathcal{PL}_{t}\): a reasoning trace paired with an ordered sequence of sub-tasks required to accomplish \(\mathcal{I}\). By hierarchically decoupling planning from execution, \(A_{p}\)—trained through a three-stage pipeline to emulate human problem-solving via skill decomposition—identifies executable, dependency-free actions and arranges them into a coherent, interpretable roadmap, such as: \texttt{\textcolor{softgreen}{1.MoveTo(printer)}}, \texttt{\textcolor{softgrey}{2.Print(e-file)}}, \texttt{\textcolor{softgrey}{3.DeliverTo(Alice)}}. The corresponding transformation is:
\begin{equation}
A_{p}:\;
(\mathcal{I},\, \mathcal{PC}_{t},\, \mathcal{RF}_{t})
\xrightarrow{\text{\tiny $A_{p}$}}
\mathcal{PL}_{t}
\end{equation}

%--------------------------------------
\subsubsection{Agent Assigner ($A_{s}$) -- Robot–Human Coordination}

\(A_{s}\) consumes the ToA plan from \(A_{p}\), the robot profiles and skill library, and active contacts/groups from \(\mathcal{PC}_t\), alongside few-shot prompts covering both robot-only and robot-human hybrid scenarios. By eliciting latent language understanding from the LLM, \(A_{s}\) produces clear assignments (e.g., \texttt{assign(MoveTo(Printer), Robot1)} or \texttt{request(Alice, Printing(loc=Printer))}), maximizing autonomous execution and deferring to humans only when robots are incapable
The agent’s input–output mapping is defined as follows:
\begin{equation}
A_{s}:\;
(\mathcal{I},\, \mathcal{PC}_{t},\, \mathcal{RF}_{t},\, \mathcal{PL}_{t},\, \mathcal{W}_{t})
\xrightarrow{\text{\tiny $A_{s}$}}
\sigma_{t}.
\end{equation}

%--------------------------------------
\subsubsection{Agent Validator ($A_{v}$) -- Pre-Execution Verification}

\(A_{v}\) uses a dedicated LLM as a judge to verify \(\sigma_t\) against \(\mathcal{I}\), \(\mathcal{W}_{t}\), and the summary of prior execution in \(\mathcal{RF}_t\), rather than relying on a continuation of the planner’s or assigner’s chain-of-thought to rethink, thereby reducing bias accumulation. \(\mathcal{V}_t\) contains (i) a validity flag {\texttt{\textcolor{softgreen}{pass}}/\texttt{\textcolor{softred}{fail}}} assessing the correctness of \(\sigma_t\), and (ii) a brief diagnostic thought explaining the verdict. If invalid, the explanation guides \(A_{p}\)’s replanning by identifying specific mismatches. This validation enhances robustness in human-interactive tasks where execution errors are costly.  
Formally, the agent computes the following transformation:
\begin{equation}
A_{v}:\;
(\mathcal{I},\, \mathcal{W}_{t},\, \mathcal{RF}_{t},\, \sigma_{t})
\xrightarrow{\text{\tiny $A_{v}$}}
\mathcal{V}_{t}
\end{equation}

\begin{figure}[!htbp]
    \centering
    \includegraphics[width=1\linewidth]{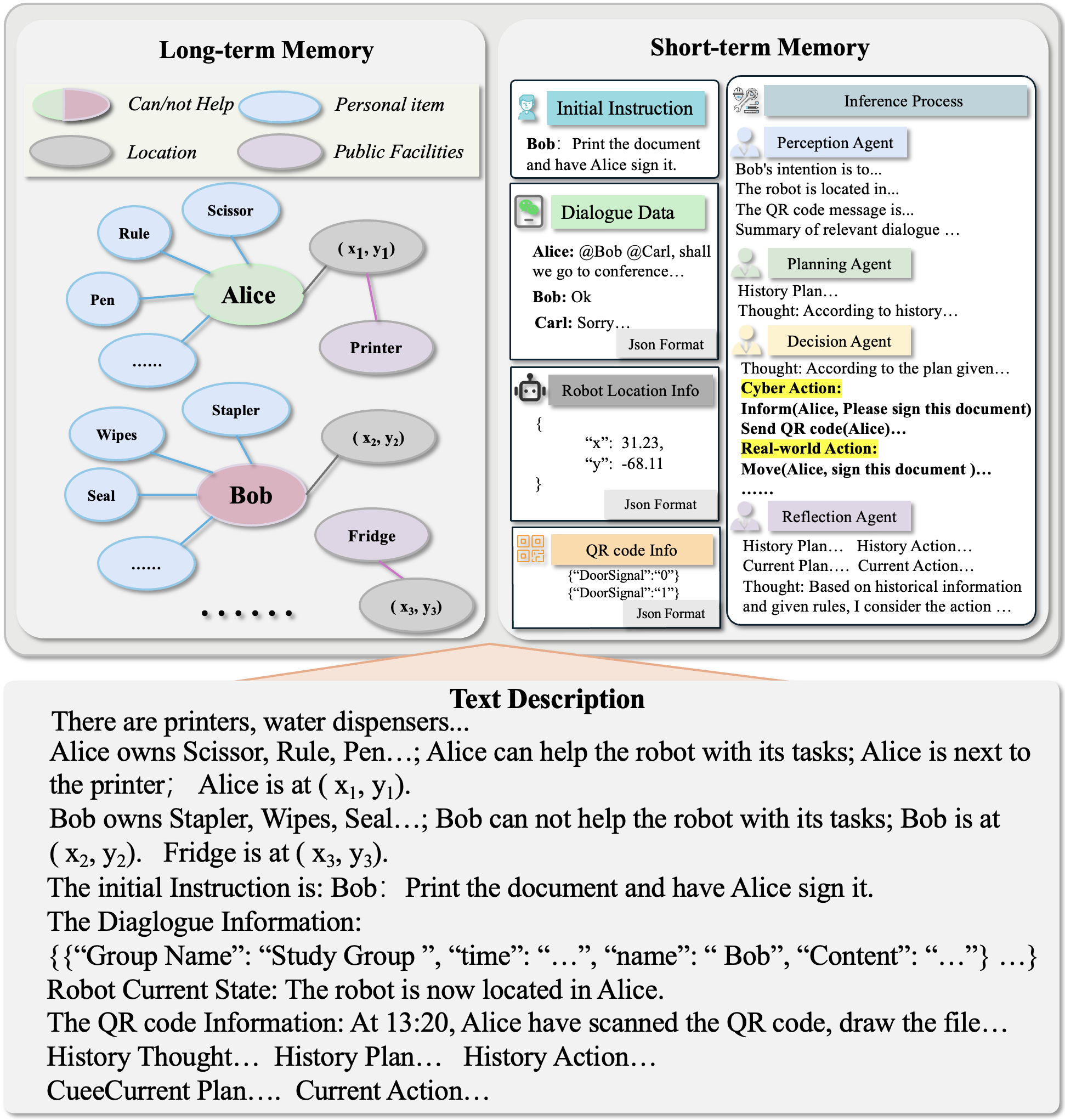}
    \vspace{-3mm}
    \caption{\textbf{Overview of the Memory Unit.} 
The left panel illustrates the long-term memory graph, which incrementally encodes cross-task knowledge, entity relations, and reusable dependencies accumulated throughout past interactions. 
The right panel shows the short-term memory, which stores task-specific observations, intermediate reasoning states, and execution feedback to support real-time planning and adaptation.
}
    \label{fig:memory unit}
\end{figure} 

\subsection{Memory Unit}
\label{sec:memory}

The Memory Unit serves as the foundational layer of the entire framework. It maintains both \emph{long-term} and \emph{short-term} memories (see Fig.~\ref{fig:memory unit}). 
Short-term memory stores the transient information generated during each end-to-end execution of a user request, including the instruction, recent dialogue context, and embodied states such as the robot’s location or the smart locker’s status. 
The agent’s textual outputs retained in the shared message pool also function as part of short-term memory, enabling rapid within-session reasoning without repeatedly invoking perception modules.

Long-term memory, in contrast, is modeled as an undirected topological graph. Nodes correspond to humans, public facilities, personal items, and physical locations, while edges encode their relationships---for example, item ownership or the spatial placement of humans and facilities. 
Human nodes additionally maintain an availability attribute indicating their readiness to participate in collaborative tasks. 
During execution, the Perceiver incrementally updates node attributes and relations to reflect environmental changes, ensuring that the memory remains concise and semantically grounded rather than a raw, ever-growing log of observations.

We denote the overall memory state at time step~$t$ as $\mathcal{M}_t$, which can be serialized into textual descriptions and supplied to the foundation models as part of their contextual input. Combined with the Perceiver’s refinement of memory entries, this mechanism reduces the token footprint, improves accuracy by filtering out irrelevant noise, and enables foundation models to perform faster and more stable reasoning for long-horizon interactive tasks.
By maintaining only on-demand, structured information, and by combining the Perceiver with this memory design, the system greatly lightens the contextual burden passed to downstream agents: the Manager, Planner, Assigner, and Validator operate on compact, semantically meaningful summaries rather than large, unfiltered histories.

\section{Dataset Construction}

\label{sec:dataset}

Constructing reliable embodied reasoning agents requires evaluation settings that reflect the complexity of real-world service robotics, where multiple humans, robots, and shared resources must be coordinated under evolving constraints. Existing benchmarks largely isolate agents from human participants or restrict interaction to static manipulation, leaving a substantial gap to practical deployment. To bridge this gap, we build a high-quality dataset integrating natural-language understanding, instruction clarification, and parallel task allocation within human-centric workflows. The dataset contains 2,100 task instances from simulated domestic and workplace environments and is organized into three tiers: 
(1) \emph{Basic flows} (300) with fully specified requests; 
(2) \emph{Ambiguous flows} (800) requiring clarification of roles, targets, or intents; 
(3) \emph{Dynamic-context flows} (1,000) involving human unavailability or resource contention that demand adaptive planning.

\subsubsection{Basic Instruction Generation}
Each task is generated from a structured JSON schema describing multi-step interactions among humans, public facilities, and personal items. Personal items are explicitly grounded to owners, while public facilities (e.g., printers, lockers) encode functional constraints. We use \texttt{DeepSeek-R1} to sample grounded entities from the environment, assigns semantic roles---\texttt{origin}, \texttt{intermediary}, and \texttt{destination} users---and generates dependency-aware action sequences forming executable service workflows, with irrelevant entities pruned automatically.

All tasks share a controlled vocabulary of 30 portable personal items and 12 public facilities, covering representative object types and functional resources commonly required in human--robot collaborative scenarios.

\begin{figure*}[t]
    \centering
    \includegraphics[width=1\linewidth]{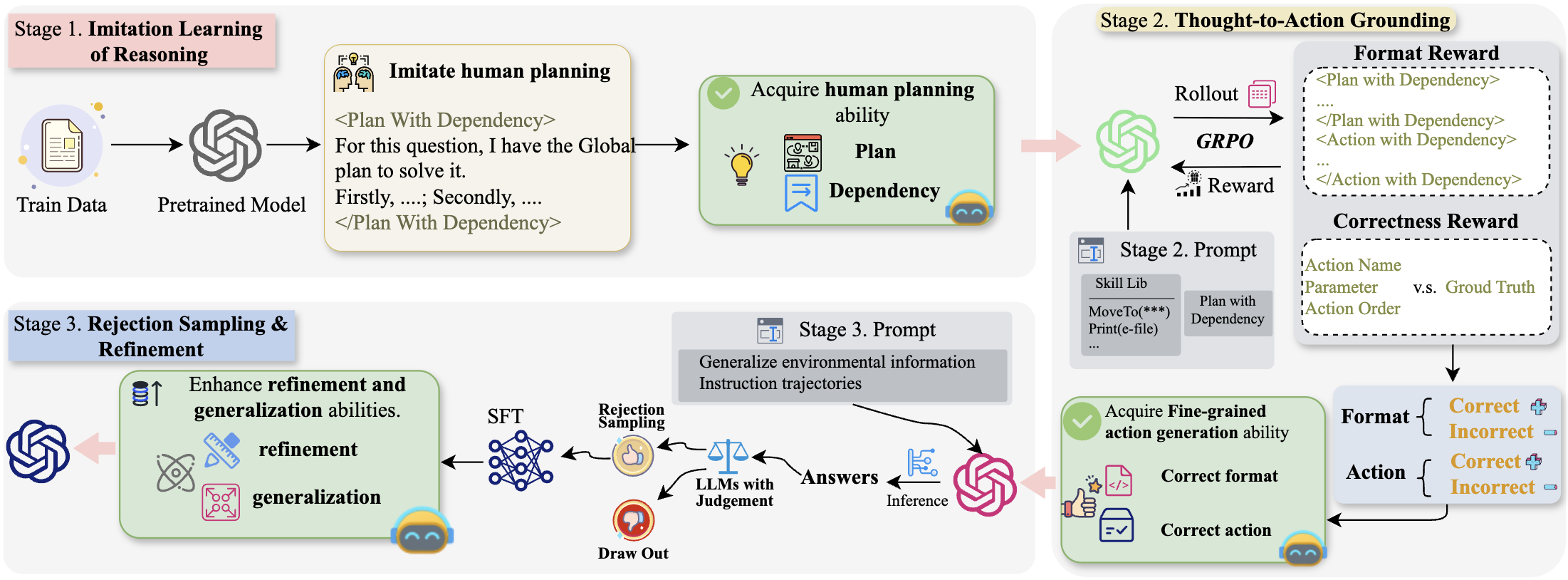}
    \vspace{-3mm}
    \caption{\textbf{The three-stage training pipeline of ToA Planning.}  
The pipeline progressively transforms natural-language instructions into structured, executable Thought-of-Action plans. \textbf{Stage~1: Imitation Learning of Reasoning.} The model learns to imitate human planning demonstrations and acquire global dependencies between reasoning steps. \textbf{Stage~2: Thought-to-Action Grounding.} Using GRPO rollouts, the model is trained to abstract fine-grained action skills, with rewards encouraging correct action formats, parameters, and ordering. \textbf{Stage~3: Rejection Sampling and Refinement.} The planner is further refined via SFT and rejection sampling, where LLM-based judges filter incorrect or low-quality outputs to enhance correctness, robustness, and generalization.}
    \label{fig:training}
    \vspace{-3mm}
\end{figure*}

\definecolor{codeblue}{rgb}{0.25,0.5,0.5}
\lstset{
  backgroundcolor=\color{gray!10},
  basicstyle=\fontsize{8pt}{9pt}\ttfamily\selectfont, % 等宽字体是关键
  columns=fullflexible,
  breaklines=True, % 禁用自动换行以保持对齐
  captionpos=b,
  commentstyle=\color{codeblue},
  showstringspaces=false,
  tabsize=20, % 用大制表位模拟表格列
  keepspaces=true, % 保留空格
  literate=
        {×}{$\times$}1   % 替换 × 为 LaTeX 的 \times
        {φ}{$\phi$}1,    % 替换 φ 为 LaTeX 的 \phi
}

\begin{lstlisting}[label=lst:objects]
Personal Item        Mass (g)            Bbox (cm)
---------------------------------------------------
Ballpoint_Pen          10                14×1×1
Stapler                220               13×4×5
Document               5               29.7×21×0.05
...

Public Facility           Purpose / Interaction
--------------------------------------------------
Printer                Print, scan, copy documents
Fridge                 Store perishable items
Coffee_Machine         Support SelectDrink
...
\end{lstlisting}

The resulting JSON structure reflects both the role-specific object associations and the dependency-aware action graph:

\definecolor{codeblue}{rgb}{0.25,0.5,0.5}
\definecolor{codekw}{rgb}{0.85, 0.18, 0.50}
\definecolor{keywordgreen}{rgb}{0,0.6,0}
\lstset{
  backgroundcolor=\color{gray!10},
  basicstyle=\fontsize{8pt}{9pt}\ttfamily\selectfont,
  columns=fullflexible,
  breaklines=true,
  captionpos=b,
  commentstyle=\fontsize{6.5pt}{7.5pt}\color{codeblue},
  keywords = {User, Assistant, System}, 
  keywordstyle = {\textbf},
  label={lst:planning_prompt}
}

\begin{lstlisting}[label=listing:basic_template, basicstyle=\fontsize{8}{9}\selectfont\ttfamily]
{
  "origin": ["person_who_issued_request"], 
  "destination": ["final_recipient"],

  "facility": ["required_facility"],
  "facility_usage_action": ["action_to_use_facility"],

  "personal_item_1": ["item_name"], 
  "personal_item_1_owner": ["item_owner"], 
  "personal_item_1_actions": [
    ["action", "person", "dependency(optional)"],
    ...
  ],
  "intermediate_person_1": ["person_involved_in_transit_or_signature"],
   ...
}
\end{lstlisting}

This structure ensures that every entity is role-assigned and every action is grounded to an actor and, when necessary, a prerequisite. Natural-language instructions are generated from the structured JSON and validated through a semi-automated pipeline: 83\% of samples pass automatic checks, and the remaining 17\% are manually corrected, with duplicates removed. The resulting set contains 300 high-quality, dependency-aware task specifications serving as the basic service flows.

\subsubsection{Ambiguity Injection}
To simulate underspecified real-world requests, we convert each basic task into an ambiguous variant by masking one critical semantic field (e.g., facility usage, personal item, required actor, or destination). Few-shot prompting with \texttt{GPT-4o} then produces an ambiguous instruction together with a corresponding clarification question, generated from both the complete and masked JSON. This yields 800 ambiguity-driven cases.

\subsubsection{Dynamic Context Simulation}
To model realistic disruptions, we augment each task with a \texttt{dynamic\_context} field specifying time-varying human availability. A subset of replaceable actors is randomly marked as unavailable; this information is hidden from the planner and used only by the simulator to trigger failures when unavailable individuals are referenced. These failures require the planner to revise its strategy, enabling evaluation under partial observability, imperfect inputs, and runtime recovery. This stage adds 1,000 dynamic-context tasks.

\section{Thought-of-Action Planning}
\label{sec:toa}

We introduce a \emph{Thought-of-Action} (ToA) representation that bridges free-form chain-of-thought reasoning and executable primitive skills. Each action is typed, parameterized, and dependency-aware, enabling parallelism and precondition checking. Compared with generic chain-of-thought, ToA provides a structured, dependency-aware representation that enables the Assigner to dispatch actions to humans or robots and allows the Validator to check feasibility before execution. This compact formulation significantly reduces token usage and improves planning efficiency.

Our planner is trained through a three-stage pipeline(Fig.~\ref{fig:training}):

\subsubsection{Stage 1: Imitation Learning of Reasoning}
We fine-tune an LLM (\texttt{Qwen3-8B} in InteractGen) on 5{,}000 supervised examples constructed from 1{,}500 tasks randomly sampled from Section~\ref{sec:dataset}. Each sample is augmented with intermediate \texttt{perception} and \texttt{reflection} states, allowing the planner to learn mid-execution reasoning rather than cold-start decomposition. Manually annotated, dependency-aware plans, which can be seen as a form of simple chain-of-thought, serve as supervision targets.

Given $\mathcal{D}=\{(x_i,y_i)\}$ where $x_i$ concatenates all context fields and $y_i$ is the plan token sequence, the IL objective is:
\begin{equation}
\mathcal{L}_{\mathrm{IL}}
=
-\frac{1}{N}
\sum_{i,t}
\log p_{\theta}(y_{i,t}\mid y_{i,<t},x_i).
\end{equation}
This establishes strong priors for coherent, dependency-aware reasoning.

\subsubsection{Stage 2 – Thought-to-Action Grounding} 

Building on \textsc{Stage~1}, we further refine the model by aligning its free-form reasoning with executable skill primitives. We introduce a human-atomic skill library across public domains, including actions like \texttt{walkto(location)}, \texttt{pickup(object)}, \texttt{print(document)}, etc. Each training sample maps an instruction to a structured \textit{Thought-of-Action} plan—a sequence of minimal skills from the library. For example, \texttt{``Alice: Print a report in the USB Drive for the meeting in 502 hosted by Bob''} is grounded as: \textcolor{softgreen}{\texttt{WalkTo(Alice)}}, \textcolor{softgrey}{\texttt{Retrieve(USB Drive)}, \texttt{WalkTo(Printer)}, \texttt{Insert(USB Drive)}, \texttt{Print(Report)}, \texttt{DeliverTo(Bob, Room 502)}}.

\paragraph{Human‑Atomic Skill Library}
We focus on \emph{human‑atomic} skills that robots may not yet master fully but that humans can readily perform. Each skill record includes following components: 

\begin{lstlisting}[label=lst:objects]

SkillName(parameters), e.g. MoveTo(Alice), 
Order: (step = n, dep = k), where dep denotes the 
most recent prerequisite step.

\end{lstlisting}

\paragraph{Dataset Augmentation for GRPO}
We create 5,000 \emph{skill‑level trajectories} by pairing every \textsc{Stage~1} context with its aligned skill sequence.  Two strategies were evaluated:
\begin{enumerate}
  \item \textit{Expert Annotation:} human annotators map the gold plan to skills—high accuracy but costly.
  \item \textit{Model Bootstrapping:} the \textsc{Stage~1} planner auto‑generates a plan, which is then heuristically parsed into skills and provides on‑policy diversity.
\end{enumerate}
Empirically we combine both: 60\% expert‑labelled seeds ensure correctness; 40\% model‑generated samples broaden coverage and expose current stage to its own prediction distribution. The data format is as follows:
\begin{lstlisting}
{ 
  "instruction": "user instruction", 
  "input": {
    "env info": "prior knowledge", 
    "perception": "perception of last action", 
    "reflection": ["Y/N", "description", "history"], 
  },
  "output": {
    "planning with dependency": "planning" 
    "action with dependency": "human-atomic skills"
  }
}
\end{lstlisting} 

\paragraph{Reward Design for GRPO.}
To optimize the planner, we adopt GRPO—a group-based variant of PPO that evaluates multiple sampled action sequences per query. Since GRPO requires a scalar reward \( r_i \) for each predicted sequence \( s_i \), we design a reward function that captures both structural validity and semantic correctness, measuring how closely each predicted skill sequence aligns with the ground-truth execution plan.

Formally, the reward evaluates four aspects: 

\textbf{Format validity.}
We first check whether the output is structurally valid---i.e., whether the required fields (planning dependencies, skill sequence, parameters) appear in the correct format:
\begin{equation}
    \mathcal{R}_{\text{format}} =
\begin{cases}
1, & \text{all fields are present and formatted} \\
0, & \text{otherwise}
\end{cases}
\end{equation}

\textbf{Skill-name matching.}
Let \( N_G \) and \( N_P \) be the sets of skill names appearing in the ground-truth sequence \( G \) and predicted sequence \( P \). Their similarity is measured using the Jaccard index:
\begin{equation}
    r_{\text{name}} = \frac{|N_G \cap N_P|}{|N_G \cup N_P|}, \quad r_{\text{name}} \in [0, 1]
\end{equation}

\textbf{Parameter matching.}
For each ground-truth skill \( G_j \), we compute the Jaccard similarity of its parameter set against the matched predicted skill \( P_j \):
\begin{equation}
\begin{aligned}
r_{\text{param}} = \frac{1}{|G|} \sum_{G_j \in G}
\frac{
    |\mathrm{params}(G_j \cap P_j)|
}{
    |\mathrm{params}(G_j \cup P_j)|
},
\quad r_{\text{param}} \in [0,1]
\end{aligned}
\end{equation}

\textbf{Ordering consistency.}
We further evaluate whether the predicted sequence preserves necessary action dependencies and subtask ordering:
\begin{equation}
\begin{aligned}
r_{\text{order}}
&=
\frac{1}{\sum_{G_j \in G} |\mathrm{orders}(G_j)|}
\sum_{G_j \in G} 
\sum_{o \in \mathrm{o}(G_j)}
\frac{1}{2}\,\mathbf{1}[P_j[o] = G_j[o]],
\\[2pt]
&\qquad\qquad r_{\text{order}} \in [0,1]
\end{aligned}
\end{equation}

\noindent \textbf{Combined matching score.}
The three semantic scores are summed to obtain an overall matching value:
\begin{equation}
    r_{\text{match}} = r_{\text{name}} + r_{\text{param}} + r_{\text{order}}, 
\qquad r_{\text{match}} \in [0, 3]
\end{equation}

We then linearly normalize this into the correctness reward:

\begin{equation}
    \mathcal{R}_{\text{correct}} = 2 \cdot r_{\text{match}} - 3, 
\qquad \mathcal{R}_{\text{correct}} \in [-3, 3]
\end{equation}

\noindent \textbf{Final reward.}
The final reward used by GRPO is:
\begin{equation}
    \mathcal{R}_{\text{final}} 
= \mathcal{R}_{\text{format}} + \mathcal{R}_{\text{correct}}, 
\qquad
\mathcal{R}_{\text{final}} \in [-3, 4]
\end{equation}

For each query \( Q \), GRPO samples a group 
\(\mathcal{G}_Q = \{A, (s_1,r_1), \dots, (s_n,r_n)\}\),  
where \( A \) is the ground-truth plan and each \( s_i \) is a predicted sequence assigned reward \( r_i \).

We compute the group mean and standard deviation:
\begin{equation}
    \mu_Q = \frac{1}{n} \sum_{i=1}^{n} r_i, \quad
\sigma_Q = 
\sqrt{\frac{1}{n} \sum_{i=1}^{n} (r_i - \mu_Q)^2}
\end{equation}

The normalized advantage of each sample is:
\begin{equation}
    A_i(s_i \mid Q) = \frac{r_i - \mu_Q}{\sigma_Q + \eta}
\end{equation}
where \( \eta \) is a small constant for numerical stability. The final GRPO optimization objective can be represented as:
\begin{equation}
\begin{aligned}
J_{\text{GRPO}}(\theta)
&=
\mathbb{E}_Q \Big[
\min \Big(
\frac{\pi_\theta(s_i \mid Q)}{\pi_{\text{old}}(s_i \mid Q)} A_i,
\\
&\qquad\qquad
\text{clip}\!\left(
\frac{\pi_\theta(s_i \mid Q)}{\pi_{\text{old}}(s_i \mid Q)},
1-\epsilon,\;1+\epsilon
\right) A_i
\Big)
\Big]
\end{aligned}
\end{equation}
where \( \pi_\theta \) and \( \pi_{\text{old}} \) denote the current and reference policies, and \( \epsilon \) controls the clipping range to prevent overly large policy updates.

Our group-normalized objective stabilizes training by reducing reward variance across queries and provides a much finer supervision signal than binary or coarse scoring. By evaluating skill names, parameters, and ordering jointly, the reward reflects the true structure of tool-using tasks, leading to plans that are more consistent and executable. In practice, this yields faster convergence and stronger performance than imitation learning alone. GRPO-trained planners produce cleaner, dependency-correct action sequences and are less prone to formatting or semantic errors, while also learning to request clarification when uncertain.

\subsubsection{Stage 3: Rejection Sampling and Refinement}
We expand the task pool with 500 additional instructions (300 curated + 200 adapted from ALFWorld~\cite{shridhar2021alfworldaligningtextembodied} and BEHAVIOR~\cite{srivastava2021behaviorbenchmarkeverydayhousehold}). From the GRPO-trained planner, we sample $K=10$ diverse ToA candidates per task. A high-capacity verifier (\texttt{Gemini~2.5 Pro}) filters candidates based on logical consistency, semantic validity, and conciseness. A total of 8{,}123 trajectories are accepted.

Each accepted trajectory $\tau=(y,a)$ supervises both planning and action decoding:
\begin{equation} \label{eq:rft_enriched} \begin{aligned} \mathcal{L}_{\mathrm{RST}}(\theta) = -\frac{1}{|\mathcal{T}_{\mathrm{acc}}|} \sum_{(y,a)\in\mathcal{T}_{\mathrm{acc}}} \Big[ &\sum_{t=1}^{|y|} \log p_\theta\!\left(y_t \mid y_{<t}, x\right) \\ &+\; \sum_{t=1}^{|a|} \log p_\theta\!\left(a_t \mid a_{<t}, y, x\right) \Big] \end{aligned} \end{equation}

This final refinement distills verifier-approved reasoning patterns, yielding a planner that is accurate, dependency-consistent, and robust in real-world workflows.

\begin{figure*}[!t]
    \centering
    \includegraphics[width=1.0\linewidth]{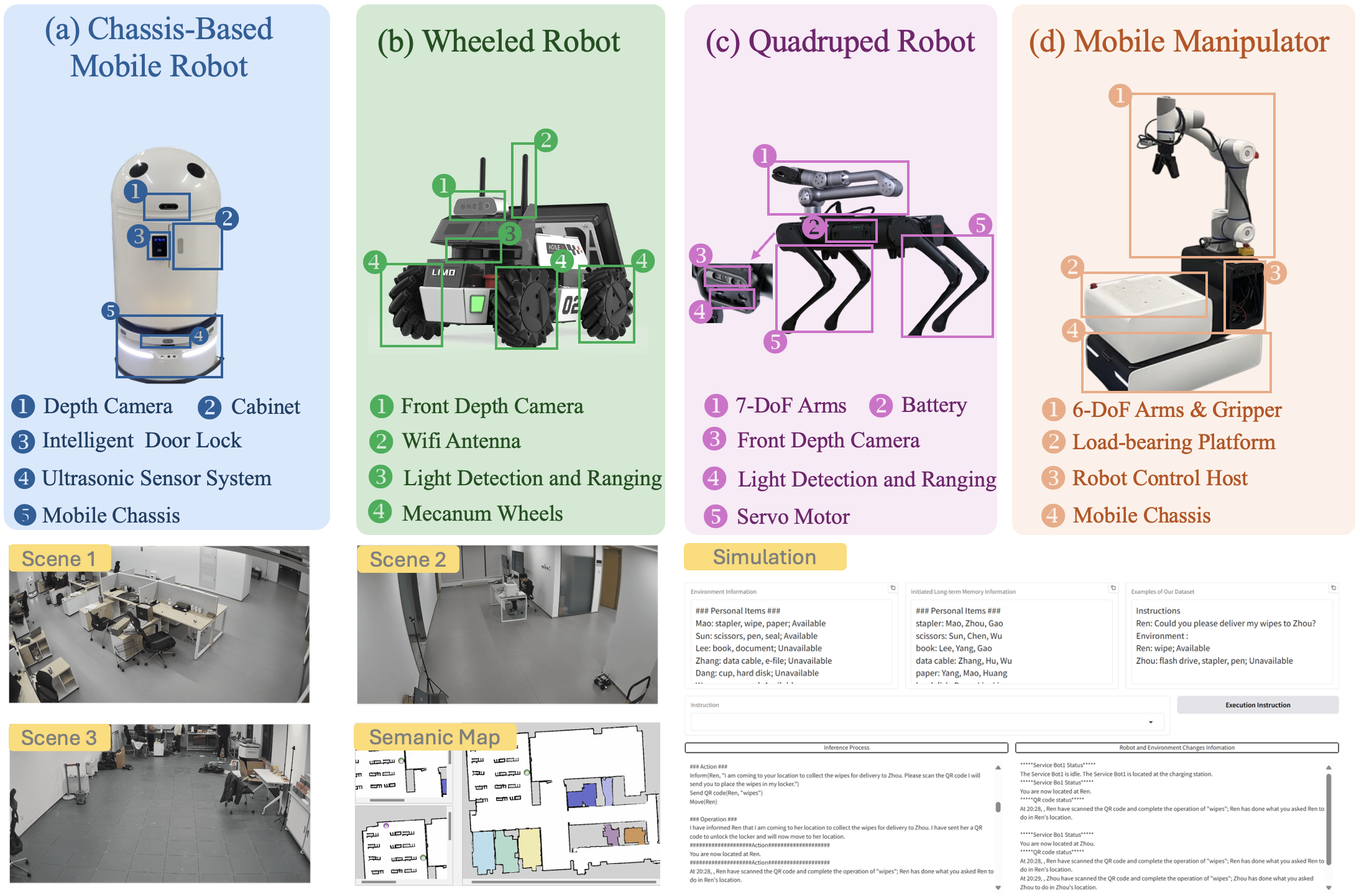}
    \vspace{-3mm}
   \caption{\textbf{Overview of the experimental setup.}  
Our system operates across a heterogeneous fleet of embodied robots and an interactive real-world environment, accompanied by a mirrored simulation environment with a language interface.  
The top row shows the four robot platforms used in our study—mobile base, wheeled robot, quadruped, and mobile manipulator—highlighting their sensing and actuation capabilities.  
The bottom row illustrates the corresponding semantic maps, real-world scenes, and simulation interface that together constitute the operational environment for executing human–robot collaborative tasks.
}
    \label{fig:real_scene}
    \vspace{-3mm}
\end{figure*}

\section{Experiment}
\label{sec:experiment}

Our evaluation targets service-oriented workflows that couple item operations, user localization, and human–robot interaction subtasks (e.g., requesting signatures, borrowing objects, performing handovers) in fluid, multi-user settings. Experiments are conducted in two parallel environments sharing identical robot profiles and coordination protocols. The simulation environment extends TextWorld~\cite{côté2019textworldlearningenvironmenttextbased} to support multi-user office scenes with symbolic robot skills and natural-language user responses. The real-world environment is a 400\,m\textsuperscript{2} lab–office (see Fig.~\ref{fig:real_scene}) containing multiple users, 12 personal-item categories, and 6 shared facilities. Evaluation spans 300 unseen simulated tasks (five runs each) and 167 real-world tasks (three runs each, including ablations), with mirrored APIs and capabilities ensuring consistent comparison.

\subsection{Robot Profiles}

We deploy a heterogeneous team of four robot types, each contributing specialized mobility and manipulation capabilities. The Assigner allocates subtasks using structured robot profiles encoding affordances, workspace limits, sensing, and dexterity.

\paragraph{Mobile Base with Smart Locker (two units)}
A chassis-based mobile robot equipped with smart lockers handle autonomous delivery. An RGB-D camera and 2D LiDAR support navigation, while QR-code sensing provides reliable feedback on locker contents. Lacking external manipulation, they specialize in secure transport.

\paragraph{Wheeled Robot with Rear Tray}
A lightweight shuttle platform carrying a passive rear tray. Basic cameras and navigation sensors support flexible intra-room transfers.

\paragraph{Quadruped Robot with Manipulator}
A legged robot with an integrated arm for operation in cluttered, uneven, or low-elevation environments. Its manipulation controller is trained via the ACT framework~\cite{zhao2023learningfinegrainedbimanualmanipulation} using 600 teleoperated demonstrations collected for tasks like pick, place, and handover. Data include synchronized RGB-D streams and end-effector poses obtained through a joystick–tablet teleoperation setup.

\paragraph{Mobile Manipulator}
A wheeled base with a 6-DOF arm serving as the team’s most dexterous platform. It excels at mid- and high-elevation tasks (desk/shelf retrieval, drawer operations, handovers). Its ACT controller is trained on 800 teleoperated trajectories gathered via a VR interface, with synchronized joint states and RGB-D data.

\paragraph{Human Agent}
Humans act as flexible collaborators for tasks requiring fine dexterity, domain knowledge, or capabilities beyond the robots. The Assigner issues symbolic requests (e.g., \texttt{ask\_human(print, file)}), and humans respond through a dedicated messaging thread. InteractGen's LLM agents run on a separate compute node and dispatch symbolic commands to robots through ROS. The Perception and Reflection agents maintain a concise, continuously updated semantic world state appended to each planning prompt.

\subsection{Real-World Environment}

Our real-world deployment takes place in a 400\,m\textsuperscript{2} lab–office environment designed to emulate realistic domestic scenarios (see Fig.~\ref{fig:real_scene}). The space is divided into functional zones (e.g., meeting room, workspace) and populated with 6 human participants, 12 personal-item categories, and 6 public facilities.

To support autonomous multi-robot planning and execution, we integrate \textit{five types of perceptors} that provide multi-modal situational awareness. These perceptors interface with physical hardware (RGB-D cameras, LiDAR, force sensors) or cloud APIs (dialogue streams), continuously assisting with updating the unified memory unit shared across all agents.

\textbf{Vision Perceptor (Detection and Scene Understanding).}  
Each mobile robot carries an Intel RealSense D435 RGB-D camera. An RGB-based detector identifies household objects, while depth maps provide 3D approximations of their positions. When finer granularity is required, an image-segmentation model extracts object masks. Detected objects are anchored to the semantic map with spatial coordinates.

\textbf{Localization and Navigation Perceptor.}
Robots localize using 2D LiDAR-based SLAM (RPLIDAR A3 with SLAMTEC SLAM), maintaining a shared $(x,y,\theta)$ frame across the heterogeneous fleet. A ROS Navigation Stack planner evaluates navigation feasibility and produces collision-free paths, allowing low-level mobility to be handled reliably and freeing our system to focus on interaction-centric reasoning for multi-user service workflows. The mobile robots combine LiDAR, wheel odometry, and an IMU to run a 2D pose-graph SLAM back-end that generates a 5\,cm occupancy grid. A semantic layer is built on top of this grid, annotating default locations of users, public facilities, and personal items through either automatic segmentation (e.g., Mask2Former) or manual annotation. The resulting metric--semantic map provides a compact set of task-relevant nodes that supports efficient reasoning about multi-user presence and navigation without requiring end-to-end vision--language navigation pipelines.

\textbf{Force/Tactile Perceptor.}  
Manipulator-equipped robots use a wrist-mounted force–torque sensor and gripper touch sensors to monitor grasping. Excessive forces or unexpected torque profiles signal grasp failure, while rapid finger-width reduction indicates successful pickup. Signals are processed at 50\,Hz through ROS. 

\textbf{Dialogue Perceptor.}
A cloud-based interface monitors message threads and uses a retrieval API to stream new instructions, clarifications, and updates in real time, enabling language-driven multi-step planning. The system also registers as a digital user in chat applications; contact and messaging APIs link each person to a cyber account whose logs provide cues about availability, attention, and likely locations. Group channels supply shared context, forming a unified dialogue corpus that augments perception without additional fusion.

\textbf{Smart-Lock Perceptor.}  
Chassis robots equipped with smart lockers use QR-code scanning and electronic lock-state feedback to confirm deposit and retrieval operations. 

\subsection{Simulation Environment}
\label{appendix:sim-env}

We build a text-based household simulator on top of TextWorld~\cite{côté2019textworldlearningenvironmenttextbased}, extended to multi-room indoor scenes with manipulable objects and multiple embodied agents, mirroring domestic and office-like service tasks. Agents (robots or humans) can be issued with structured high-level commands (e.g., \texttt{navigate to <location>}, \texttt{pick up <object>}, \texttt{ask <human> for help}) and maintain an internal inventory of held items. To ensure interpretability, the simulator adopts deterministic symbolic actions rather than free-form language. All commands are parsed into templated feedback similar to ALFWorld~\cite{shridhar2021alfworldaligningtextembodied}, providing either confirmations for valid actions or standardized messages for unmet preconditions or incorrect contexts, enabling reliable multi-agent training and evaluation.

Robots are instantiated using the same configuration profiles as in Fig.~\ref{fig:real_scene}. Each simulated robot model is governed by payload, workspace, and access constraints, and annotated with its available low-level skills. Attempts to execute actions beyond a robot’s capability deterministically fail, encouraging collaborative behaviors such as dynamic delegation and resource-aware planning. Multi-agent interactions, including object handovers, are explicitly modeled: a successful transfer requires an \texttt{offer} followed by a matching \texttt{take}, reflecting realistic turn-based coordination.

When a robot issues a help request (e.g., \texttt{ask human to lend <object>}), the simulator uses an LLM-driven conversational policy conditioned on task context. If the context specifies that a person is unavailable, the environment returns rejection or silence (e.g., “Sorry, I’m busy right now.”), capturing uncertainty in human-in-the-loop workflows. Although initial human information is provided as prior knowledge, availability may shift mid-task, requiring adaptive reasoning and replanning. This design provides a controlled but realistic foundation that supports consistent multi-agent evaluation while approximating key dynamics observed in the real-world system.

\subsection{Baseline}
\label{sec:baselines}
We compare InteractGen with two types of complementary baselines with metrics shown in TABLE~\ref{tab:metrics}:  
\textbf{Generic Agentic Reasoning Paradigms}—\textit{ReAct}~\cite{yao2023react}, \textit{Reflexion}~\cite{shinn2023reflexionlanguageagentsverbal}, and \textit{Mobile-Agent-v2}~\cite{wang2024mobileagentv2mobiledeviceoperation} represent task-agnostic frameworks that interleave chain-of-thought reasoning with action execution or self-repair; although not tailored to embodied datasets, they supply strong multi-agent planning heuristics; 
\textbf{LLM-Powered Multi-Robot Controllers}—\textit{CoELA}~\cite{zhang2024buildingcooperativeembodiedagents}, \textit{CaPo}~\cite{liu2025capocooperativeplanoptimization}, \textit{RoCo}~\cite{mandi2023rocodialecticmultirobotcollaboration}, \textit{SMART-LLM}~\cite{kannan2024smart}, \textit{Lip-LLM}~\cite{obata2024lipllmintegratinglinearprogramming}, \textit{EMOS}~\cite{chen2025emosembodimentawareheterogeneousmultirobot}, \textit{LaMMA-P}~\cite{zhang2025lammapgeneralizablemultiagentlonghorizon}, and \textit{HMCF}~\cite{li2025hmcfhumanintheloopmultirobotcollaboration}, engineered for embodied tasks, exposing declarative skill APIs to dispatch heterogeneous robots thus serving as representative ``LLM for multi-robot'' baselines:

\begin{table}[!t]
\centering
\caption{Evaluation Metrics}
\label{tab:metrics}
\renewcommand{\arraystretch}{1.25}

\begin{tabularx}{0.95\linewidth}{l X}
\toprule
\textbf{SR} & \textit{Success Rate}-- Percentage of fully completed instructions, allowing redundant steps \\
\midrule
\textbf{CR} & \textit{Completion Rate}-- Proportion of expected interaction steps completed \\
\midrule
\textbf{RR} & \textit{Redundancy Rate}-- Fraction of redundant actions among successful ones \\
\midrule
\textbf{RSR} & \textit{Robot Subtask Rate}-- Proportion of subtasks completed by the robot out of all assignments \\
\midrule
\textbf{TOKEN} & \textit{Token Usage}-- Average tokens used per successful execution, measured by the tokenizer \\
\midrule
\textbf{TIME} & \textit{Execution Time}-- Total wall-clock time (in seconds) from task reception to completion \\
\bottomrule
\end{tabularx}
\vspace{-5mm}
\end{table}

\begin{itemize}[leftmargin=1.5em]
\item \textbf{ReAct}~\cite{yao2023react}.  
      Designed for single-agent tool use, ReAct interleaves chain-of-thought and actions but offers no primitives for robot selection or human queries.  
      We instantiated \emph{one} ReAct agent controlling all four robots sequentially and exposed a single ``\texttt{ask human}'' action.  
      The agent seldom invoked this action, so ambiguities stayed unresolved; moreover, sequential control prevented true parallelism, bottlenecking multi-robot throughput.

\item \textbf{Reflexion}~\cite{shinn2023reflexionlanguageagentsverbal}.  
      Adds self-critique to ReAct.  
      While this fixed minor execution slips, reflections occur \emph{post hoc}; the agent still cannot clarify intent in advance, wasting steps before correcting.  
      Parallel allocation issues persist because only one robot is active per reasoning step.

\item \textbf{Mobile-Agent-v2}~\cite{wang2024mobileagentv2mobiledeviceoperation}.  
      Originally maps GUI actions; we provide robot skills and augment GUI perception with text world summaries.  
      The agent handles dynamic UI but assumes a single ``device'', so concurrent robot scheduling degrades to serial execution; it barely requests human help and mis-allocates tasks requiring parallel execution.

\item \textbf{CoELA}~\cite{zhang2024buildingcooperativeembodiedagents}.  
      Accepts declarative skill APIs; we supplied four robot profiles and wrote wrappers that block skills a robot cannot perform.  
      CoELA plans centrally but, lacking a clarification loop, silently proceeds when the goal is ambiguous, sometimes assigning unreachable subtasks. Without explicit concurrency control, task decomposition tends to under-utilise robots.

\item \textbf{CaPo}~\cite{liu2025capocooperativeplanoptimization}.  
      Uses two LLM agents to co-optimise a plan; it can \emph{in principle} ask the user, yet in practice the agents debate internally and rarely query the human.  
      Divergent meta-plans stall progress under ambiguity; alternatively, both agents off-load the entire task to the human. Even with clear goals, its synchronous dialogue slows real-time execution and limits parallelism.

\begin{table*}[!t]
\centering
\small
\caption{\textbf{Quantitative comparison on our evaluation against baselines}. Arrows indicate directionality of better performance. Rates are reported as percentages (mean$\pm$std).}
\label{tab:method_comparison}
\renewcommand{\arraystretch}{1.15}
\setlength{\tabcolsep}{4pt}
\begin{tabular}{l|cc|cc|cc|cc|c|c}
\toprule
\multirow{2}{*}{Method}
& \multicolumn{2}{c|}{SR [\%] $\uparrow$}
& \multicolumn{2}{c|}{CR [\%] $\uparrow$}
& \multicolumn{2}{c|}{RR [\%] $\downarrow$}
& \multicolumn{2}{c|}{RSR [\%] $\uparrow$}
& TOKEN ($\times 10^3$) $\downarrow$
& TIME [s] $\downarrow$ \\
\cline{2-11}
& Sim & Real & Sim & Real & Sim & Real & Sim & Real & Real & Real \\
\midrule
\rowcolor[gray]{0.9}
\multicolumn{11}{c}{\textbf{Generic Methods}} \\
\midrule
ReAct~\cite{yao2023react}  
& 31$\pm$4 & 13$\pm$5
& 39$\pm$3 & 21$\pm$5
& 16$\pm$3 & 21$\pm$5
& 53$\pm$3 & 50$\pm$4
& 53.2$\pm$4.1 & 2309$\pm$119
\\
Reflexion~\cite{shinn2023reflexionlanguageagentsverbal} 
& 36$\pm$3 & 19$\pm$5
& 43$\pm$2 & 27$\pm$4
& 14$\pm$3 & 19$\pm$4
& 54$\pm$3 & 50$\pm$5
& 51.2$\pm$4.1 & 2101$\pm$111
\\
Mobile-Agent-v2~\cite{wang2024mobileagentv2mobiledeviceoperation}  
& \textbf{54$\pm$3} & \textbf{43$\pm$3}
& \textbf{54$\pm$5} & \textbf{49$\pm$3}
& \textbf{8$\pm$3}  & \textbf{13$\pm$4}
& \textbf{65$\pm$4} & \textbf{54$\pm$5}
& \textbf{43.2$\pm$3.0} & \textbf{1746$\pm$90}
\\
\midrule
\rowcolor[gray]{0.9}
\multicolumn{11}{c}{\textbf{Multi-Robot Methods}} \\
\midrule
SMART-LLM~\cite{kannan2024smart}
& 47$\pm$2 & 43$\pm$4
& 53$\pm$3 & 45$\pm$4
& 17$\pm$5 & 20$\pm$1
& 59$\pm$3 & 44$\pm$3
& 49.4$\pm$3.7 & 1996$\pm$99
\\
CoELA~\cite{zhang2024buildingcooperativeembodiedagents} 
& 44$\pm$1 & 29$\pm$4
& 52$\pm$3 & 41$\pm$6
& 19$\pm$2 & 23$\pm$3
& 52$\pm$2 & 42$\pm$1
& 51.2$\pm$4.1 & 2101$\pm$111
\\
Lip-LLM~\cite{obata2024lipllmintegratinglinearprogramming}
& 53$\pm$2 & 45$\pm$2
& 57$\pm$4 & 48$\pm$4
& 16$\pm$4 & 19$\pm$5
& 60$\pm$5 & 49$\pm$1
& 47.1$\pm$3.4 & 1746$\pm$93
\\
RoCo~\cite{mandi2023rocodialecticmultirobotcollaboration}
& 50$\pm$5 & 39$\pm$4
& 53$\pm$2 & 45$\pm$1
& 17$\pm$5 & 20$\pm$3
& 55$\pm$4 & 45$\pm$3
& 48.7$\pm$3.4 & 1823$\pm$95
\\
LaMMA-P~\cite{zhang2025lammapgeneralizablemultiagentlonghorizon}
& 56$\pm$5 & 47$\pm$7
& 62$\pm$4 & 50$\pm$6
& 12$\pm$5 & 13$\pm$6
& 62$\pm$5 & 52$\pm$5
& 44.1$\pm$3.1 & 1597$\pm$89
\\
HMCF~\cite{li2025hmcfhumanintheloopmultirobotcollaboration}
& 57$\pm$5 & \textbf{51$\pm$1}
& 66$\pm$5 & 53$\pm$4
& 8$\pm$3  & 13$\pm$5
& 62$\pm$2 & 50$\pm$1
& 43.7$\pm$2.9 & 1504$\pm$88
\\
EMOS~\cite{chen2025emosembodimentawareheterogeneousmultirobot}
& 52$\pm$5 & 48$\pm$3
& 67$\pm$4 & \textbf{54$\pm$2}
& 12$\pm$4 & 14$\pm$3
& \textbf{65$\pm$2} & 53$\pm$4
& 45.7$\pm$3.1 & 1674$\pm$89
\\
CaPo~\cite{liu2025capocooperativeplanoptimization}
& \textbf{62$\pm$4} & 50$\pm$2
& \textbf{69$\pm$5} & 53$\pm$4
& \textbf{7$\pm$3}  & \textbf{11$\pm$5}
& 63$\pm$4 & \textbf{55$\pm$2}
& \textbf{41.7$\pm$2.8} & \textbf{1367$\pm$88}
\\
\midrule
\rowcolor[gray]{0.9}
\multicolumn{11}{c}{\textbf{InteractGen with Different Planning Foundation Models}} \\
\midrule
InteractGen (Planner on GPT-4o)  
& 68$\pm$3 & 57$\pm$4 
& 72$\pm$4 & 63$\pm$3
& 6$\pm$3  & \textbf{\underline{5$\pm$2}}
& 66$\pm$3 & 62$\pm$2
& 36.1$\pm$1.9 & 1137$\pm$65
\\
InteractGen (Planner on DS-R1)   
& 74$\pm$3 & 59$\pm$5 
& 75$\pm$3 & 63$\pm$3
& 6$\pm$3  & 7$\pm$3
& 78$\pm$3 & 63$\pm$4
& 34.7$\pm$1.7 & 1061$\pm$63
\\
InteractGen (ToA Planner)   
& \textbf{\underline{77$\pm$3}} & \textbf{\underline{70$\pm$3}} 
& \textbf{\underline{80$\pm$5}} & \textbf{\underline{74$\pm$4}}
& \textbf{\underline{3$\pm$0}}  & 6$\pm$1
& \textbf{\underline{81$\pm$3}} & \textbf{\underline{68$\pm$5}}
& \textbf{\underline{31.3$\pm$1.3}} & \textbf{\underline{970$\pm$56}}
\\
\bottomrule
\end{tabular}
\vspace{-3mm}
\end{table*}

\item \textbf{RoCo}~\cite{mandi2023rocodialecticmultirobotcollaboration}.  
      Employs dialectic negotiation among robot-side LLMs, which rarely issue human queries—revealing a robot--robot rather than robot--human focus.  
      Its distributed reasoning is latency-heavy: independent LLMs produce inconsistencies, leading to conflicting allocations.

\item \textbf{SMART-LLM}~\cite{kannan2024smart}.  
      Translates tasks into linear temporal logic for assignment.  
      Although prompts encourage clarification, the agent seldom asks; ambiguity breaks the LTL translator.  
      Dynamic human availability invalidates pre-solved plans, and it lacks a replanning trigger.

\item \textbf{LiP-LLM}~\cite{obata2024lipllmintegratinglinearprogramming}.  
      Couples an LLM with a linear-programming solver to optimise resource use, but no native clarification means wrong assumptions propagate.  
      The solver also ignores action duration, so simultaneous allocation overloads individual robots.

\item \textbf{EMOS}~\cite{chen2025emosembodimentawareheterogeneousmultirobot}.  
      Leverages structured ``robot resumes'' for reasoning but assumes fully specified task goals. It lacks language grounding to convert natural instructions into concrete predicates, often defaulting to arbitrary parameters. Once a plan is generated, it cannot be updated mid-execution—EMOS offers no support for human feedback or replanning under dynamic changes.

\item \textbf{LaMMA-P}~\cite{zhang2025lammapgeneralizablemultiagentlonghorizon}.  
      Integrates an LLM with a PDDL planner.  
      We provided a handcrafted PDDL domain. Replanning is possible but slow—every environmental change triggers full PDDL reconstruction.

\item \textbf{HMCF}~\cite{li2025hmcfhumanintheloopmultirobotcollaboration}.  
      Positions a human overseer who may intervene.  
      We added a single ``\texttt{ask human for help}'' primitive; the LLM controller rarely used it, causing the human to intervene only \emph{after} failures, increasing recovery cost.  
      HMCF distributes one LLM per robot; concurrent reasoning raises bandwidth and consistency issues, making the overall coordination process a bottleneck.
\end{itemize}

To ensure a fair comparison, all baselines above are instantiated with the same foundation model, \texttt{GPT-4o}, following their original architectural design and prompting logic. 

In InteractGen, the Perceiver and Assigner likewise uses \texttt{GPT-4o}, while the Verifier and Manager are implemented with \texttt{DeepSeek-R1}. Our Planner is trained in three stages:  

\textbf{\textsc{Stage~1:} Supervised Fine-Tuning.}  
Full-parameter SFT with DeepSpeed ZeRO-2 on eight A100 (\texttt{bf16}, 2k context). Per-GPU batch size 1 with 4 accumulation steps (effective batch size 32); learning rate \(2\times10^{-5}\); trained for 3 epochs. 

\textbf{\textsc{Stage~2:} GRPO.}  
Trained with ZeRO-2 under the same batch/precision settings for 5 epochs at \(1\times10^{-6}\). Each query samples 4 candidate plans to form GRPO groups. The KL penalty is removed to accelerate learning. 

\textbf{\textsc{Stage~3:} Rejection Sampling Fine-Tuning.}  
Full-parameter training resumes with ZeRO-3 on eight GPUs (\texttt{bf16}, 2k context). Per-GPU batch size 2 with 4 accumulation steps (effective batch size 64); learning rate \(2\times10^{-5}\); trained for 4 epochs. Rejection samples are generated using \texttt{vLLM} on a single A800 (\texttt{temperature=1}, \texttt{top\_p=0.7}); invalid generations are discarded. The final planner is deployed under \texttt{vLLM} with \texttt{temperature=0.2}, \texttt{top\_p=0.95}.

For the Manager’s clarification module, we use 700 ambiguous instructions from Section~\ref{sec:dataset} (600 train / 100 val) in the format \textit{instruction → split JSON of task → masked key → clarification}. We fine-tune \texttt{LLaMA-3.1-8B-Instruct} using LoRA on a single RTX\,4090 (24\,GB) for 10 epochs (\texttt{bf16}, 1024-token context). Inference uses deterministic decoding (\texttt{temperature=0}). Baseline models rely on prompt engineering to produce clarification when needed.

\section{Results}
\label{sec:results}

\begin{figure*}[!htbp]
    \centering
    \includegraphics[width=1.0\linewidth]{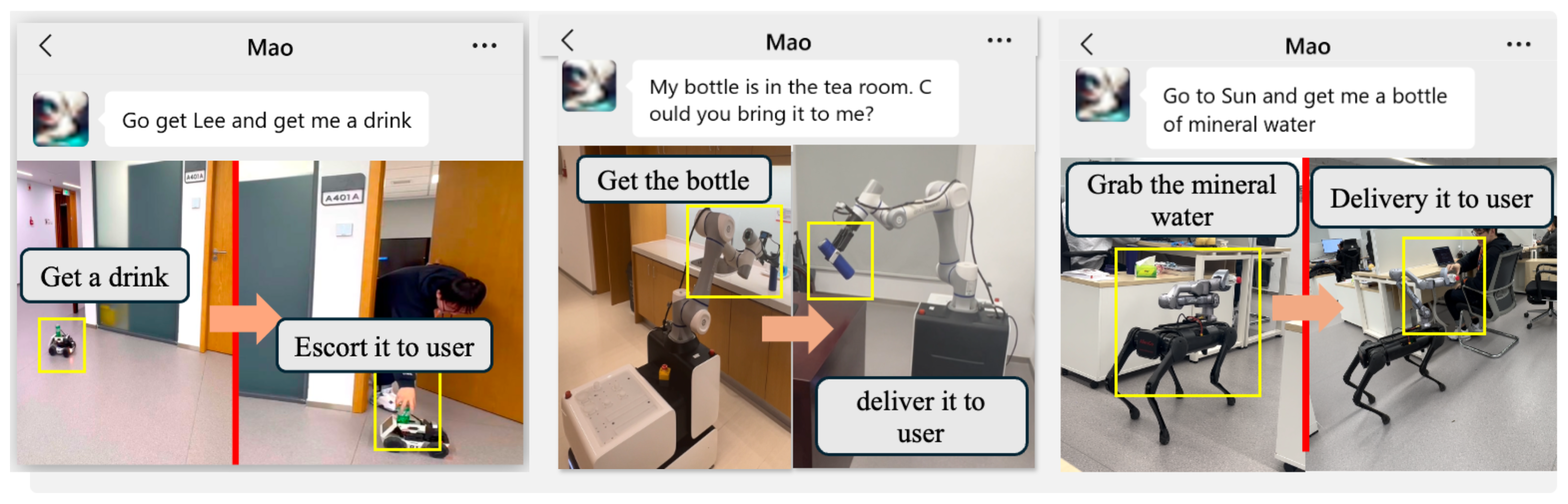}
    \vspace{-4mm}
\caption{\textbf{Examples of multi-robot service execution in real-world open-use deployment.} 
InteractGen interprets natural-language requests from users and automatically assigns appropriate robot platforms to execute the required subtasks. 
Left: a mobile robot is instructed to fetch a drink and escort it to the user. 
Middle: a mobile manipulator retrieves a bottle from the tea room and delivers it to the requester. 
Right: a quadruped robot picks up mineral water from a designated location and brings it to the user, demonstrating adaptive task allocation across heterogeneous robots and dynamic human–robot environments.}

    \label{fig:deployment_2}
\end{figure*}

In this section, we present the experimental results of InteractGen on multi-user, service tasks (see Fig.~\ref{fig:deployment_2}). We report comparative results against existing LLM-powered multi-agent frameworks and multi-robot systems, analyze the contribution of individual agent modules through ablations, and assess efficiency, usability, and user experience in real-world deployment. 

\begin{figure*}[!htbp]
    \centering
    \includegraphics[width=1.0\linewidth]{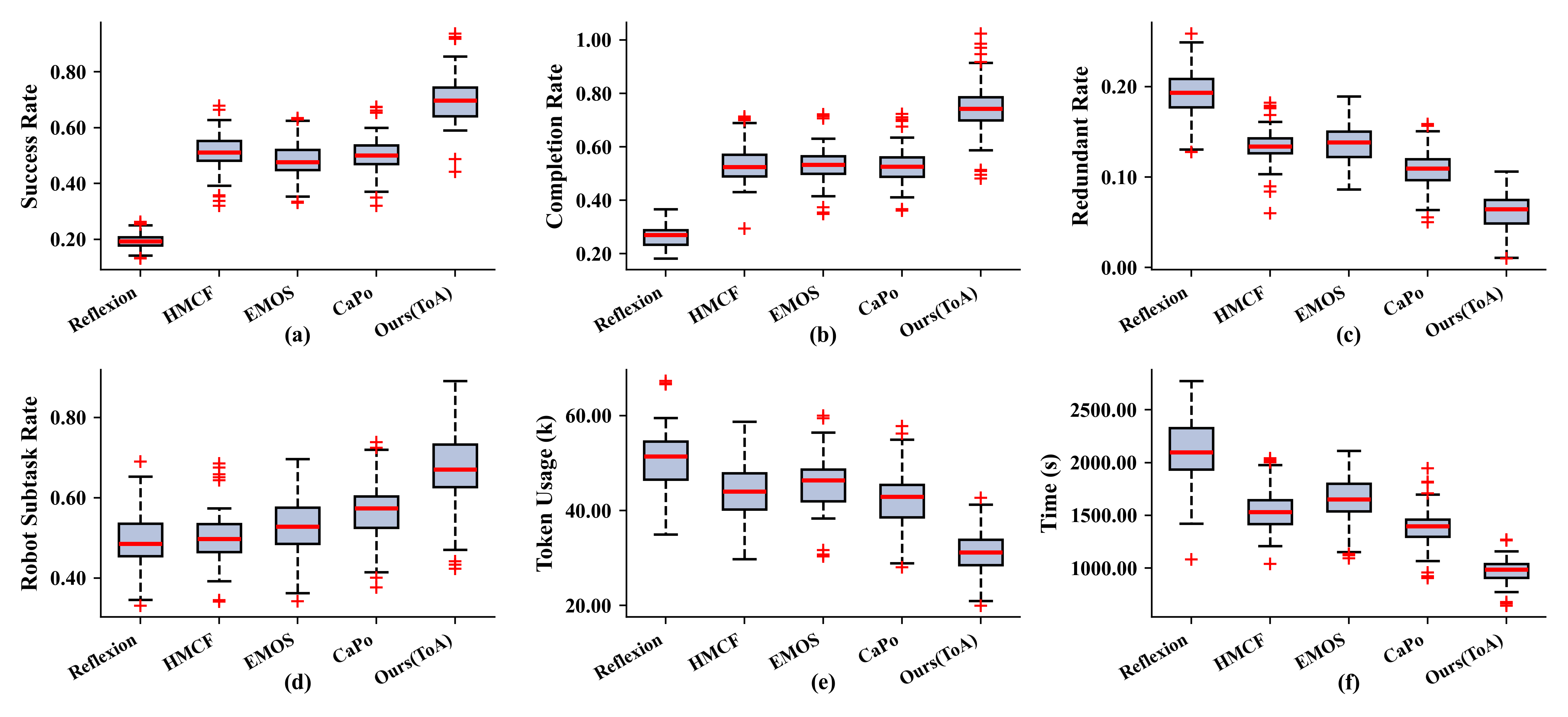}
    \vspace{-6mm}
    \caption{\textbf{Real-world performance comparison across six evaluation metrics.}  
Box plots report (a) Success Rate, (b) Completion Rate, (c) Redundant Rate, (d) Robot Subtask Rate, (e) Token Usage, and (f) Execution Time for five representative methods: Reflexion, HMCF, EMOS, CaPo, and InteractGen (ToA). InteractGen achieves the highest SR, CR, and RSR while maintaining the lowest redundancy and resource cost, as reflected by reduced token usage and shorter execution time. Baseline methods exhibit lower median performance and larger variance across metrics, highlighting InteractGen’s superior consistency, efficiency, and robustness in real-world human–robot interaction tasks.
    }
    \label{fig:compare}
    \vspace{-3mm}
\end{figure*}

\begin{figure*}[!htbp]
    \centering
    \includegraphics[width=1.0\linewidth]{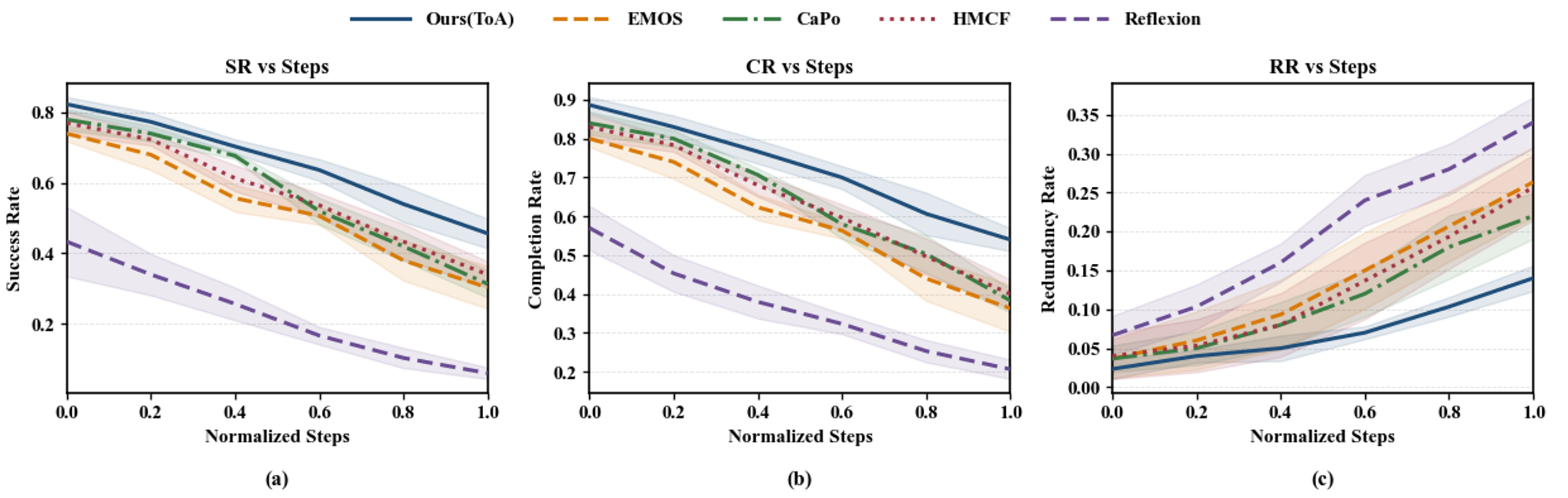}
    \vspace{-5mm}
    \caption{\textbf{Performance over normalized task steps.}  
Task steps are normalized by the number of required human interactions and object-related operations in each task, providing a unified measure of progress across scenarios.  
\textbf{(a)} Success rate, \textbf{(b)} completion rate, and \textbf{(c)} redundancy rate as tasks advance.  
Our ToA-based method sustains higher success and completion rates and consistently lower redundancy throughout the entire execution horizon, whereas baseline methods degrade more rapidly as the number of required interactions increases.} 
    \label{fig:line}
\end{figure*}

\begin{figure*}[!htbp]
    \centering
    \includegraphics[width=1.0\linewidth]{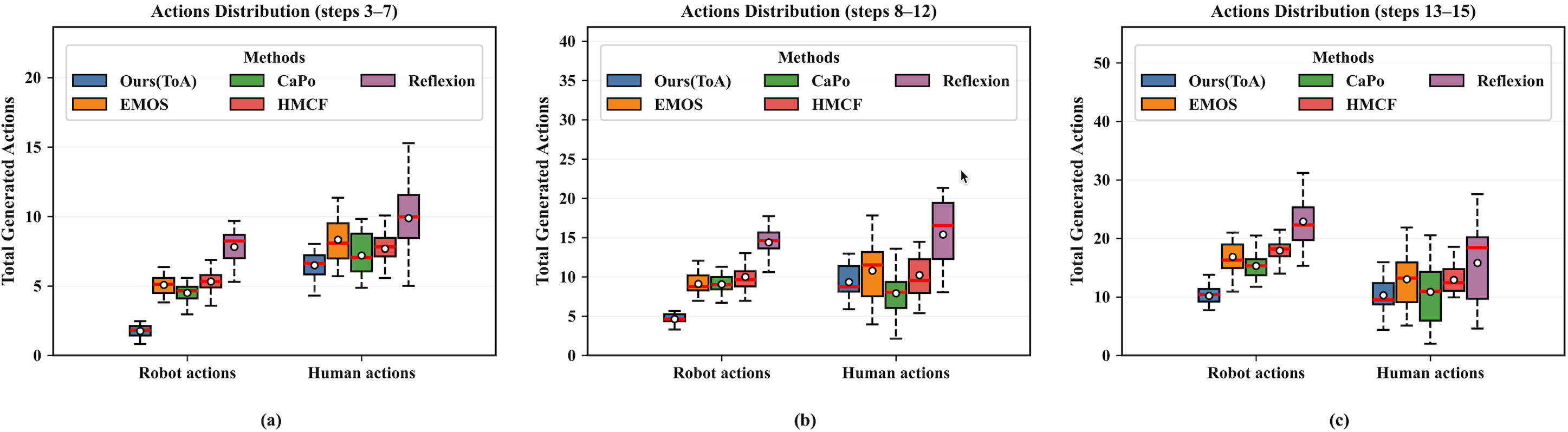}
    \vspace{-4mm}
    \caption{\textbf{Action distribution across different interaction stages.}  
Box plots show the number of robot actions (left) and human actions (right) generated by each method at three intervals: \textbf{(a)} steps 3--7, \textbf{(b)} steps 8--12, and \textbf{(c)} steps 13--15. InteractGen (ToA) consistently produces fewer and more focused robot and human action requests compared with CaPo, EMOS, HMCF, and Reflexion, indicating more efficient reasoning and reduced unnecessary agent calls. As tasks progress to later stages, baseline methods exhibit increasing variance and action redundancy, whereas InteractGen maintains stable and concise action generation.
    }
    \vspace{-3mm}
    \label{fig:rsr}
\end{figure*}

\subsection{Performance on Robot Coordination and Human Interaction}
\label{sec:performance}

\paragraph{InteractGen consistently outperforms all baselines.}
As shown in Table~\ref{tab:method_comparison}, \textit{InteractGen} with the \textit{ToA Planner} achieves the best overall performance. In simulation, it reaches the highest Success Rate (SR = \textbf{0.77}) and Completion Rate (CR = \textbf{0.80}), outperforming the strongest baseline, \textit{CaPo} (SR = 0.62, CR = 0.69), by over 15\%. Redundancy Rate (RR) drops to \textbf{0.03}, and Robot Subtask Rate (RSR) rises to \textbf{0.81}. Similar trends are observed in the real world (SR = \textbf{0.70}, CR = \textbf{0.74}). These gains stem from ToA’s modular and interpretable design, where planning, assignment, and validation are explicitly structured, tightly coordinated, and grounded in fine-grained task dependencies. While planning with \texttt{GPT-4o} lowers RR (0.05 vs.\ ToA’s 0.06), its flatter reasoning often misses dependencies, whereas ToA achieves better global consistency through explicit, fine-grained subtask structuring. \textit{InteractGen} consistently maintains a high success rate while requiring remarkably minimal human assistance, even as the complexity of human–object interactions substantially increases (see Fig.~\ref{fig:compare} and Fig.~\ref{fig:line} for more intuitive visual analyses).

\paragraph{Structured ToA Planner improves robustness beyond general-purpose LLMs.}
ToA’s advantage also comes from its dedicated training pipeline for long-horizon, multi-agent planning. Unlike general-purpose LLMs like \texttt{GPT-4o}, ToA encodes structured reasoning patterns that improve task decomposition and allocation. This leads to fewer planning errors and better resource utilization, as reflected in its top RSR (\textbf{0.81}) and lowest RR (\textbf{0.03}). It even outperforms the deep-thinking model \texttt{DeepSeek-R1}, demonstrating that careful structural design can surpass raw reasoning depth alone. These results confirm that \textit{Thought-of-Action} planning yields more reliable behavior than generic LLM prompting, offering clearer task decomposition and stronger execution consistency (see Fig.~\ref{fig:rsr}).

\paragraph{Decentralized methods exhibit lower reliability and weak human-robot coordination.}
Baselines like \textit{SMART-LLM}, \textit{HMCF}, and \textit{CaPo} rely on decentralized multi-LLM setups, where each robot runs its own planner. This often leads to redundant or conflicting behaviors, especially under partial observability or limited communication. For instance, \textit{HMCF} reports SR = 0.57 and RR = 0.08 in simulation, while \textit{CaPo} shows CR = 0.69 but suffers from poor human subtask allocation (RSR = 0.63). Without a dedicated centralized plan backbone, agents frequently act sequentially or largely rely on human agent, reducing overall task success and parallelism.

\begin{figure*}[!htbp]
    \centering
    \includegraphics[width=1.0\linewidth]{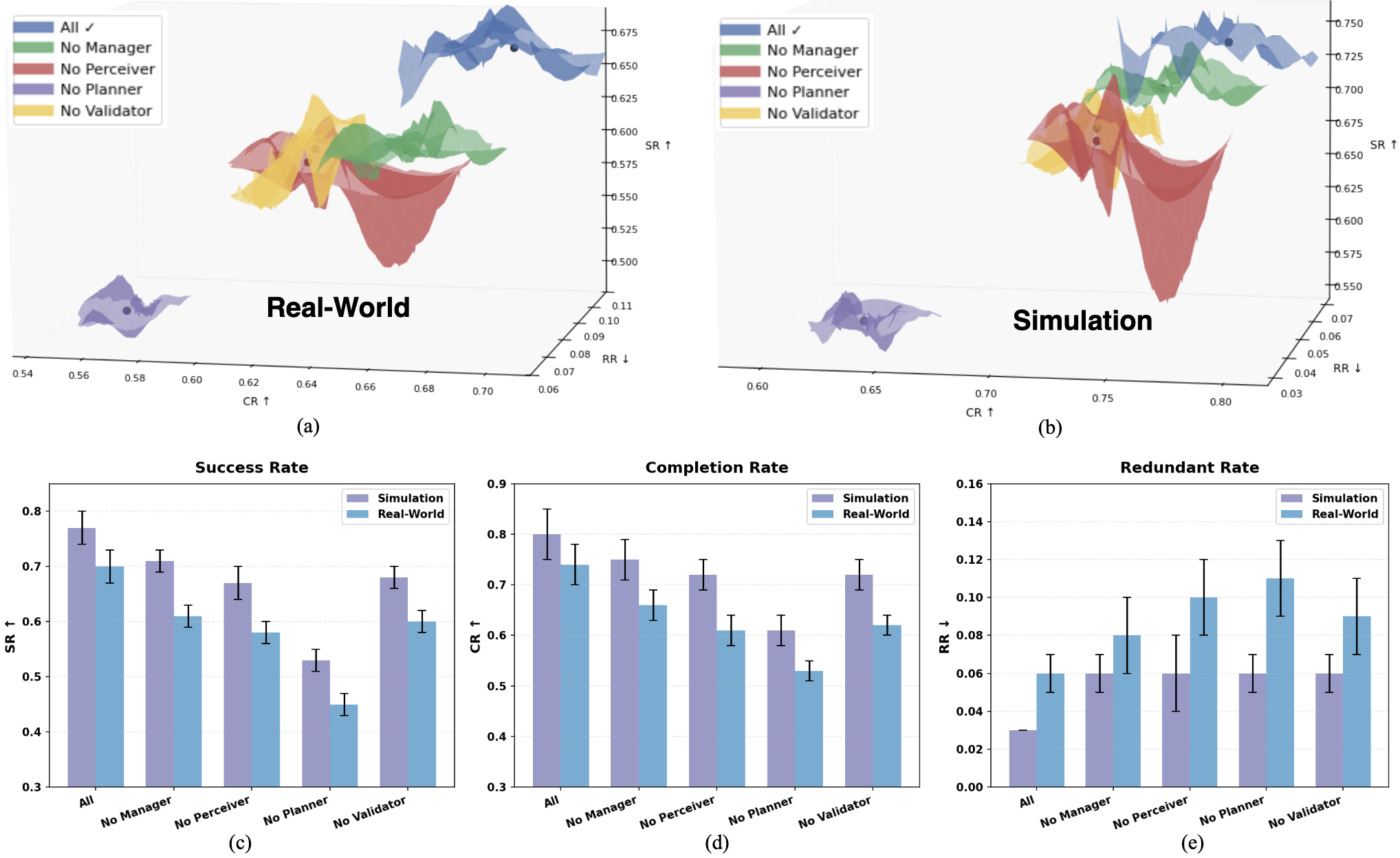}
    \vspace{-5mm}
     \caption{
   \textbf{Ablation analysis of InteractGen.} 
(a–b) 3D visualizations of Success Rate (SR), Completion Rate (CR), and Redundant Rate (RR) under module ablations in both simulation and real-world settings, highlighting the performance degradation caused by removing key agents. 
(c–e) Corresponding bar plots of SR, CR, and RR, showing consistent trends across environments and revealing each agent function as a critical component for reliable task execution.
}
    \label{fig:ablations}
\end{figure*}

\begin{figure*}[!htbp]
    \centering
    \includegraphics[width=1\linewidth]{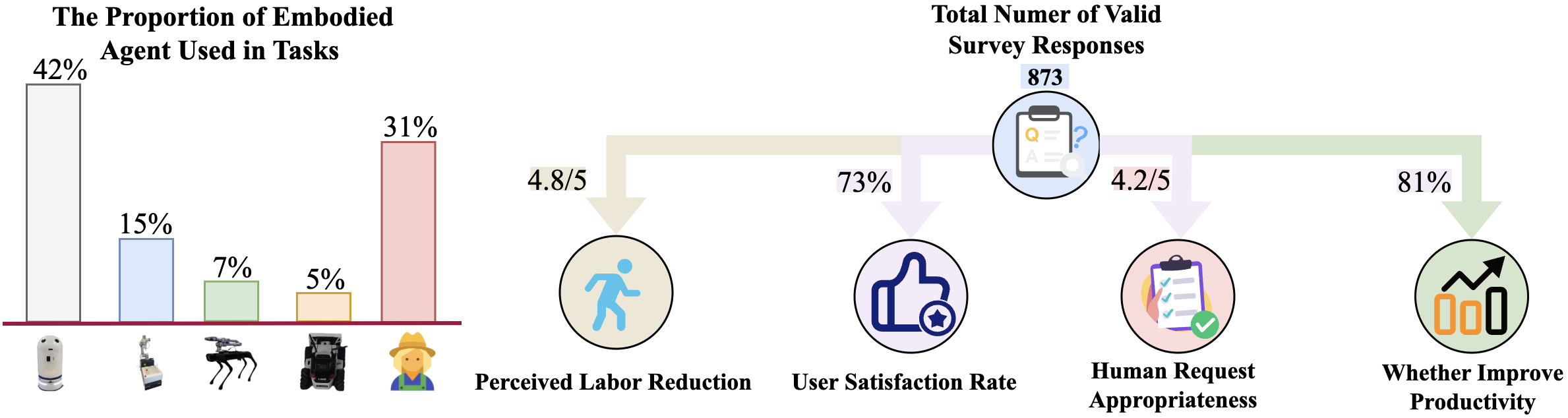}
    \vspace{-3mm}
\caption{\textbf{User survey results from a multi-month real-world deployment.} 
Left: the distribution of embodied agents involved in tasks shows that InteractGen effectively coordinates heterogeneous robots while keeping human participation lightweight. 
Right: aggregated user feedback indicates clear reductions in perceived labor, high satisfaction with system behavior, appropriate levels of human–agent requests, and noticeable improvements in overall productivity. 
These findings demonstrate InteractGen’s strong usability and its balanced integration of automation and human collaboration in practical service environments.}
    \label{fig:usersurvey}
    \vspace{-3mm}
\end{figure*}

\subsection{Token and Time Efficiency Analysis}

Despite coordinating five specialized agents, \textit{InteractGen} achieves the highest token efficiency among all compared systems. It uses only 31.3k tokens per successful run—about 40\% fewer than single-agent baselines such as Reflexion (51.2k) and ReAct (53.2k), and 25--35\% fewer than decentralized controllers including CaPo (41.7k), HMCF (43.7k), and Lip-LLM (47.1k). This efficiency stems from three design choices: (i) the Perceiver emits only incremental state updates rather than replaying full histories; (ii) the Planner produces typed, dependency-aware Thought-of-Action graphs rather than verbose free-form CoT; and (iii) all agents communicate via concise, role-specific prompts that avoid repeated grounding and unnecessary self-dialogue. These mechanisms curb token growth while preserving interpretability, enabling graceful scaling with task horizon and team size.

These design principles also yield strong time efficiency. InteractGen completes tasks in 970\,s on average, outperforming the fastest baseline (CaPo, 1367\,s) by 29\% and ReAct (2309\,s) by over 55\%. Centralized planning and assignment allow independent subtasks to run in parallel, eliminating the pairwise negotiation overhead that causes near-quadratic message growth in decentralized methods. The Validator prevents time-consuming execution of infeasible plans, and the Manager resolves ambiguities early, reducing downstream failure recovery. Ablations reinforce these observations: removing the Perceiver inflates token usage, removing the Validator prolongs fail--retry cascades, and removing the Manager introduces redundant clarifications. These results show that InteractGen’s modular pipeline is not only more robust than monolithic reasoning but also computationally lean, offering both high success rates and practical efficiency for real-world human–robot collaboration.

\begin{figure*}[!t]
    \centering
    \includegraphics[width=1.0\linewidth]{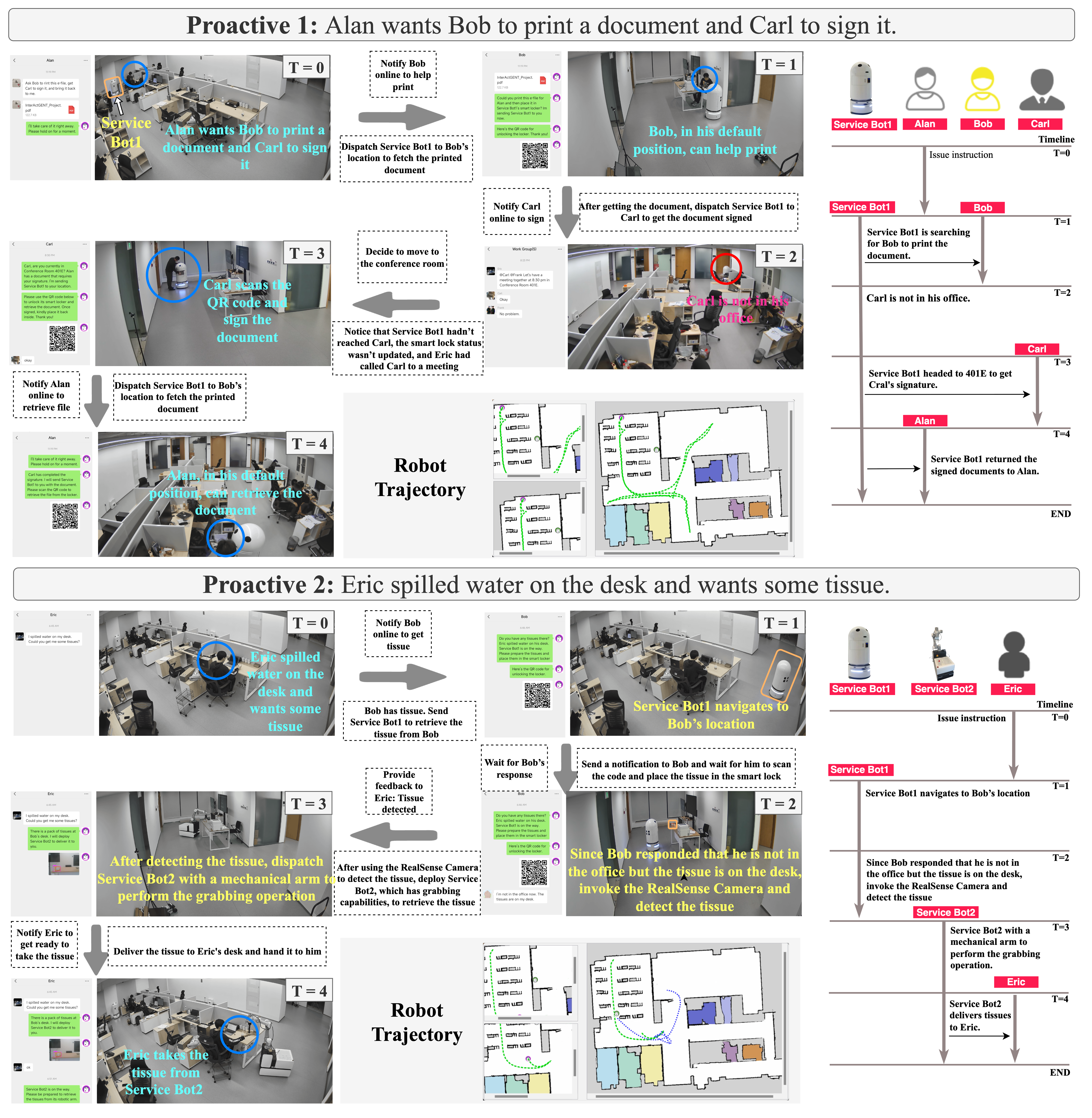}
    \vspace{-3mm}
\caption{\textbf{Proactive1:} The system adapts to human-context changes before failure occurs. Originally planning to deliver a document to Carl’s office, it detects inconsistencies—Service Bot1 not meeting Carl, unused smart lock, and a group message showing Carl in Room~401E. The Manager infers Carl has moved, triggering re-planning; the Planner redirects the robot to 401E, where Carl signs the document. This case shows proactive tracking of human movement and assumption revision. \textbf{Proactive2:} Eric requests tissue after spilling water. The system identifies Bob as a source and dispatches Service Bot1, but Bob reports he is off-site and left the tissue on his desk. The system then uses RealSense to confirm the tissue’s location and dispatches Service Bot2 (with a manipulator) to grasp and deliver it. Throughout, \textit{InteractGen} updates Eric, adapts plans from feedback, and performs the task efficiently—demonstrating robust sensing, coordination, and human-aware planning.}
    \label{fig:pro}
    \vspace{-3mm}
\end{figure*}

\subsection{Ablation Studies on Contributions of Individual Agents}
\label{sec:ablation}

We assess the functional role of each agent by removing one module at a time and replacing its behavior with a minimal prompt attached to the Assigner. This forces the system to operate without the corresponding capability and exposes its necessity. As shown in Fig.~\ref{fig:ablations}, performance degrades consistently under every ablation, demonstrating that the five-agent decomposition is structurally indispensable.

The Planner is the most critical component: removing structured task decomposition causes the success rate (SR) to fall from 0.68 to 0.45, while repetition rate (RR) nearly doubles, reflecting incoherent and looping plans typical of monolithic reasoning. Removing the Perceiver also leads to large drops (SR: 0.58), as the system loses incremental world-state updates and fails to react to dynamic changes such as user relocation. Ablating the Validator reduces SR to 0.60 because unsafe or infeasible actions bypass pre-execution checks, resulting in avoidable failures. Removing the Manager limits adaptability (SR: 0.61); without reflection and recovery mechanisms, the system cannot correct downstream errors.

These ablations verify that each LLM agent contributes a distinct and irreplaceable capability. The resulting architecture is modular, interpretable, and far more stable than a monolithic chain-of-thought pipeline. Even when compared against stronger proprietary baselines such as DeepSeek-R1, the fully assembled InteractGen system achieves superior robustness and execution quality, underscoring the practical necessity of modular specialization. We showcase two cases in Fig.~\ref{fig:pro}, demonstrating how \textit{InteractGen} executes real-world tasks through active adaptation and proactive querying, leveraging self-reflection to coordinate robots and humans.

\subsection{Real-World Performance Assessment}

We conducted a three-month open-use study (Fig~\ref{fig:usersurvey}) in a shared academic office equipped with our heterogeneous robot fleet and human participants accessible via a chat interface. Users issued natural-language requests which \textit{InteractGen} parsed and executed through its multi-agent pipeline. Subtasks were allocated dynamically: mobile chassis robots handled secure deliveries, the quadruped manipulator managed unstructured navigation and pickups, and humans were involved only for socially sensitive or dexterous operations. Across 873 valid survey responses, users reported a 73\% satisfaction rate, strong perceived labor reduction (4.8/5), and high comfort with system-issued human requests (4.2/5). Furthermore, 81\% agreed that \textit{InteractGen} improved their productivity, indicating substantial real-world impact beyond technical feasibility.

Operational statistics further show how the system coordinated mixed teams effectively. Humans performed 31\% of subtasks—mainly fine-motor or judgment-heavy actions—while mobile robots completed 42\% and manipulators 22\%, dominated by secure deliveries and simple fetch-and-carry tasks. Other platforms were used less frequently, matching the natural distribution of office demands. This division of labor highlights how \textit{InteractGen} balances autonomy with inclusivity: robots absorb routine and mobility-intensive work, while humans are engaged only when their unique strengths are essential.

Qualitative feedback emphasized reduced physical strain, lower cognitive load in coordinating shared resources, and improved fairness in delegation, as system-generated requests were perceived as neutral and equitable. These findings demonstrate that \textit{InteractGen} is not only efficient and reliable, but also socially adaptive—reinforcing its suitability as a deployable, human-centered assistant in everyday work environments.

\section{Discussion}
\label{sec:discussion}

\subsection{The Potential for LLM Agents in Embodied Tasks}

LLMs have progressed from static predictors to interactive, tool-using multi-agent systems. Yet most advances remain confined to digital settings, leaving open whether such architectures can meaningfully support embodied agents operating in dynamic, human-centered service workflows.

\begin{figure}[!t]
    \centering
    \includegraphics[width=\linewidth]{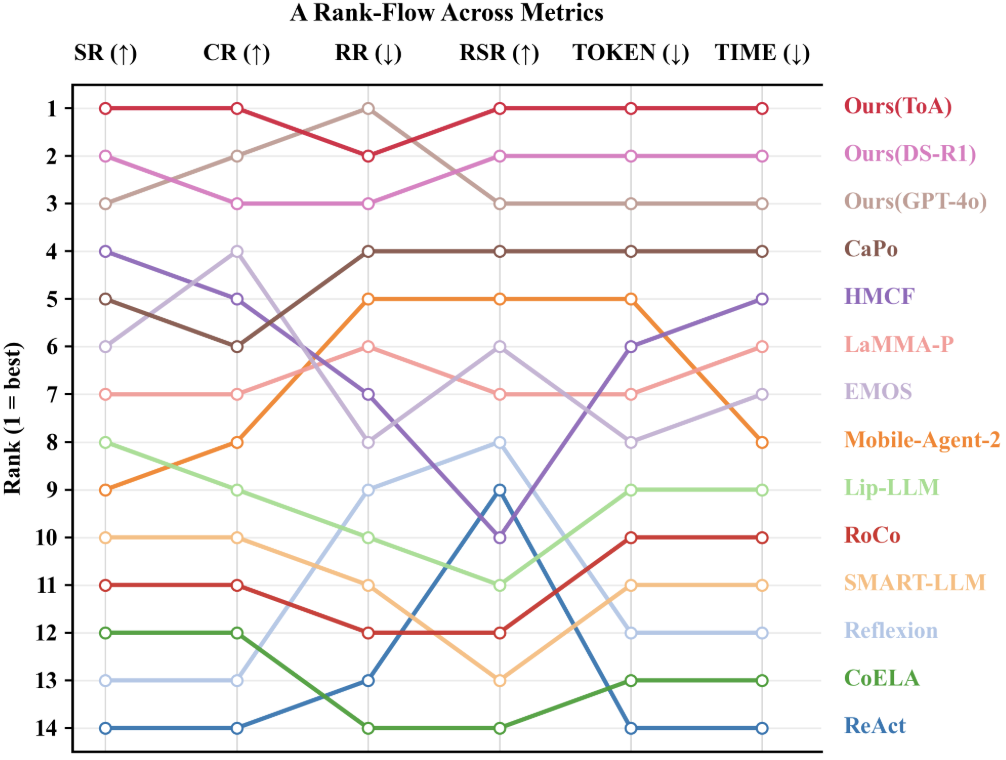}
    \caption{\textbf{Method rankings across six metrics.} 
    InteractGen variants remain near the top with minimal fluctuation, demonstrating stable advantages across all dimensions. 
    In contrast, existing baselines vary significantly across metrics, revealing trade-offs or weaknesses.}
    \label{fig:rank}
\end{figure}

Our study shows that these principles transfer naturally to embodied contexts. By organizing \textit{InteractGen} as a five-agent pipeline for perception, planning, decision, validation, and reflection, we enabled heterogeneous robots and humans to coordinate effectively in office-service routines. This modular decomposition improves interpretability and, crucially, treats humans as deployable agents rather than passive overseers. Across extensive experiments, this architecture consistently surpassed related baselines in success rate, adaptability, latency, and token efficiency, demonstrating that the collaborative strengths of digital multi-agent frameworks yield concrete performance gains in embodied workflows (see Fig.~\ref{fig:rank}).

A key advantage lies in robustness through structured delegation. In deployment, when robots encountered limitations—such as fragile object handling or user locations that were not precisely known—the reflection module dynamically reassigned subtasks to available humans or re-routed the workflow to maintain continuity. This pattern highlights a broader capability: LLM-based agent architectures can integrate diverse skills, embed human agency into the control loop, and sustain resilience in dynamic environments. We contend that such design principles represent a meaningful step toward embodied systems that are both technically competent and socially adaptive, narrowing the gap between automation and real-world human–robot collaboration.

\subsection{Foundation Models for Constructing Specialized Agents}

To enable robust human–robot mixed-team coordination, we design a comprehensive multi-agent reasoning pipeline spanning perception, planning, decision, validation, and reflection. Each stage is mapped directly to a specialized LLM agent, mirroring the cognitive workflow of a human assistant: extracting intent, perceiving the environment, generating a roadmap, selecting immediate actions, validating feasibility, requesting human help when needed, and reflecting on outcomes. This decomposition provides modularity and interpretability, while also establishing a foundation for treating humans as integral, deployable agents rather than passive overseers.

For instantiating these roles, foundation models such as GPT~\cite{openai2024gpt4technicalreport}, DeepSeek~\cite{deepseekai2025deepseekv3technicalreport}, and Gemini~\cite{geminiteam2025geminifamilyhighlycapable} series offer state-of-the-art generalization and multimodal capacity. However, when domain-specific data is available, smaller open-weight models—including \texttt{LLaMA3.1-8B-Instruct}, \texttt{Qwen3-8B}, and \texttt{Mistral-8B-Instruct}—can be fine-tuned for planning and decision stages, yielding strong performance at lower cost. In contrast, perception, validation, and reflection stages typically benefit from large-scale pretrained models due to their richer contextual grounding. Careful prompt specialization~\cite{ye2024promptengineeringpromptengineer} and targeted few-shot examples~\cite{codaforno2023metaincontextlearninglargelanguage} further enhance reasoning quality and consistency across agents.

Our experiments validate this design. Table~\ref{tab:morecomparison} shows that fine-tuned open-weight models can already outperform proprietary baselines after \textsc{Stage~2}, with \texttt{Qwen3-8B} achieving the strongest gains. Its success rate (SR) and completion rate (CR) improved from 0.25/0.53 at \textsc{Stage~1} to \textbf{0.77}/\textbf{0.80} at \textsc{Stage~3}, also yielding the lowest replan rate (RR=0.03). \texttt{LLaMA3.1-8B} and \texttt{Mistral-8B} likewise approached \texttt{Claude-3.7-Sonnet} performance by \textsc{Stage~3}. These results highlight two key insights: (i) multi-stage fine-tuning with role specialization is crucial for embodied multi-agent systems, and (ii) well-tuned open-weight models can rival or surpass proprietary alternatives, making deployment more scalable and cost-effective.

\subsection{Scalable Coordination and Generalization in InteractGen}

Single-agent paradigms such as ReAct~\cite{yao2023react} and Reflexion~\cite{shinn2023reflexionlanguageagentsverbal} entangle perception, planning, and reflection into a monolithic chain of thought, often yielding incomplete reasoning and inefficient execution. Decentralized multi-robot systems like CoELA~\cite{zhang2024buildingcooperativeembodiedagents} and RoCo~\cite{mandi2023rocodialecticmultirobotcollaboration} distribute reasoning but suffer from dialogue-induced latency that scales poorly with agent count. \textit{InteractGen} provides a middle ground through a modular five-agent pipeline in which each agent performs independent reasoning and exchanges only concise, role-specific messages. This design delivers both scalability and interpretability, enabling efficient execution of complex service workflows in multi-human, multi-robot environments.

Modular specialization brings robustness and adaptive recovery. The Validator functions as a safety gate for filtering infeasible or unsafe outputs, while the Perceiver curbs context inflation by maintaining incremental world states rather than replaying full histories. Removing these modules produces sharp performance drops, underscoring their necessity. Combined with real-time perception and LLM-as-judge reflection, InteractGen supports a reliable \emph{act → fail → reflect → replan} loop that detects anomalies, triggers replanning, and reassigns tasks—including invoking human assistance when robots reach intrinsic limits. Fragile or thin objects (e.g., documents) are automatically routed to humans, and repeated robotic failures on unfamiliar items lead to empirical adjustments in future task allocation. This reflective cycle allows the system to evolve beyond static rules and embed humans as integral agents in the collaboration loop.

Both simulation and real-world evaluations confirm the scalability and generalization enabled by this architecture. During the evaluation process, InteractGen more than doubled the success and completion rates of ReAct-style baselines. The three-stage ToA planner (SFT $\rightarrow$ GRPO $\rightarrow$ Rejection Sampling) outperformed significantly larger proprietary models while remaining more cost-efficient. During three months of open deployment, the system handled unscripted tasks such as document signing and food delivery, with surveys reporting 73\% user satisfaction and 81\% productivity improvement across 873 responses. These results establish InteractGen as a robust, scalable, and practically deployable framework for dynamic human–robot collaboration.

% 新增一个颜色避免与 myblue 冲突
\definecolor{mycyan}{RGB}{180,220,250}
\definecolor{mylavender}{RGB}{250,245,255}
\begin{table}[!t]
\centering
\caption{Evaluation in the simulation environment of different models as planner foundation models. 
The Qwen3-8B model, after three-stage training, achieves the best performance and is used as the ToA planner in InteractGen.}

\renewcommand{\arraystretch}{1.2}
\label{tab:morecomparison}

\begin{tabular}{cc|cccc}
\toprule
\multicolumn{2}{c|}{Model} & SR$\uparrow$ & CR$\uparrow$ & RR$\downarrow$ & RSR$\uparrow$ \\
\midrule

\rowcolor[gray]{0.9} 
\multicolumn{6}{c}{\textit{Proprietary}} \\

\multicolumn{2}{c|}{\cellcolor{mygreen!40} Claude-3.7-Sonnet-thinking} 
& \cellcolor{mygreen!40} 75 
& \cellcolor{mygreen!40} 77
& \cellcolor{mygreen!40} 7
& \cellcolor{mygreen!40} 71 \\

\multicolumn{2}{c|}{\cellcolor{mygreen!40} GPT-o3-mini} 
& \cellcolor{mygreen!40} 75
& \cellcolor{mygreen!40} 79
& \cellcolor{mygreen!40} 7
& \cellcolor{mygreen!40} 73 \\

\multicolumn{2}{c|}{\cellcolor{mygreen!40} GPT-4o} 
& \cellcolor{mygreen!40} 68
& \cellcolor{mygreen!40} 72
& \cellcolor{mygreen!40} 6
& \cellcolor{mygreen!40} 66 \\

\multicolumn{2}{c|}{\cellcolor{mygreen!40} GPT-o1-mini} 
& \cellcolor{mygreen!40} 70
& \cellcolor{mygreen!40} 73
& \cellcolor{mygreen!40} 5
& \cellcolor{mygreen!40} 68 \\

\midrule
\rowcolor[gray]{0.9} 
\multicolumn{6}{c}{\textit{Open-weight}} \\

\multicolumn{2}{c|}{\cellcolor{mylavender} DeepSeek-R1} 
& \cellcolor{mylavender} 74 
& \cellcolor{mylavender} 75
& \cellcolor{mylavender} 6
& \cellcolor{mylavender} 78 \\

\multicolumn{2}{c|}{\cellcolor{mylavender} LLaMA3.1-8B-Instruct}
& \cellcolor{mylavender} 51 
& \cellcolor{mylavender} 60
& \cellcolor{mylavender} 23
& \cellcolor{mylavender} 53 \\

\multicolumn{2}{c|}{\cellcolor{mylavender} Ministral-8B-Instruct}
& \cellcolor{mylavender} 54 
& \cellcolor{mylavender} 63
& \cellcolor{mylavender} 20
& \cellcolor{mylavender} 59 \\

\multicolumn{2}{c|}{\cellcolor{mylavender} Qwen3-8B}
& \cellcolor{mylavender} 60 
& \cellcolor{mylavender} 66
& \cellcolor{mylavender} 14
& \cellcolor{mylavender} 64 \\

\midrule
\rowcolor[gray]{0.9}
\multicolumn{6}{c}{\textit{Finetuned}} \\

% Stage 1
\cellcolor{myred!10} & \cellcolor{myred!10} LLaMA3.1-8B-Instruct
& \cellcolor{myred!10} 60 & \cellcolor{myred!10} 66
& \cellcolor{myred!10} 18 & \cellcolor{myred!10} 59 \\

\cellcolor{myred!10} \textsc{Stage 1} & \cellcolor{myred!10} Ministral-8B-Instruct
& \cellcolor{myred!10} 63 & \cellcolor{myred!10} 67
& \cellcolor{myred!10} 16 & \cellcolor{myred!10} 59 \\

\cellcolor{myred!10} & \cellcolor{myred!10} Qwen3-8B
& \cellcolor{myred!10} 68 & \cellcolor{myred!10} 70
& \cellcolor{myred!10} 10 & \cellcolor{myred!10} 67 \\

% Stage 2
\cellcolor{myyellow!30} & \cellcolor{myyellow!30} LLaMA3.1-8B-Instruct
& \cellcolor{myyellow!30} 66 & \cellcolor{myyellow!30} 70
& \cellcolor{myyellow!30} 8 & \cellcolor{myyellow!30} 63 \\

\cellcolor{myyellow!30} \textsc{Stage 2} & \cellcolor{myyellow!30} Ministral-8B-Instruct
& \cellcolor{myyellow!30} 69 & \cellcolor{myyellow!30} 73
& \cellcolor{myyellow!30} 8 & \cellcolor{myyellow!30} 68 \\

\cellcolor{myyellow!30} & \cellcolor{myyellow!30} Qwen3-8B
& \cellcolor{myyellow!30} 73 & \cellcolor{myyellow!30} 75
& \cellcolor{myyellow!30} 5 & \cellcolor{myyellow!30} 74 \\

% Stage 3
\cellcolor{myblue!40} & \cellcolor{myblue!40} LLaMA3.1-8B-Instruct
& \cellcolor{myblue!40} 71 & \cellcolor{myblue!40} 74
& \cellcolor{myblue!40} 7 & \cellcolor{myblue!40} 73 \\

\cellcolor{myblue!40} \textsc{Stage 3} & \cellcolor{myblue!40} Ministral-8B-Instruct
& \cellcolor{myblue!40} 73 & \cellcolor{myblue!40} 77
& \cellcolor{myblue!40} 6 & \cellcolor{myblue!40} 76 \\

\cellcolor{myblue!40} & \cellcolor{myblue!40} Qwen3-8B
& \cellcolor{myblue!40} \textbf{77} 
& \cellcolor{myblue!40} \textbf{80}
& \cellcolor{myblue!40} \textbf{3}
& \cellcolor{myblue!40} \textbf{81} \\

\bottomrule
\end{tabular}
\end{table}

\begin{figure*}[!htbp]
    \centering
    \includegraphics[width=1.0\linewidth]{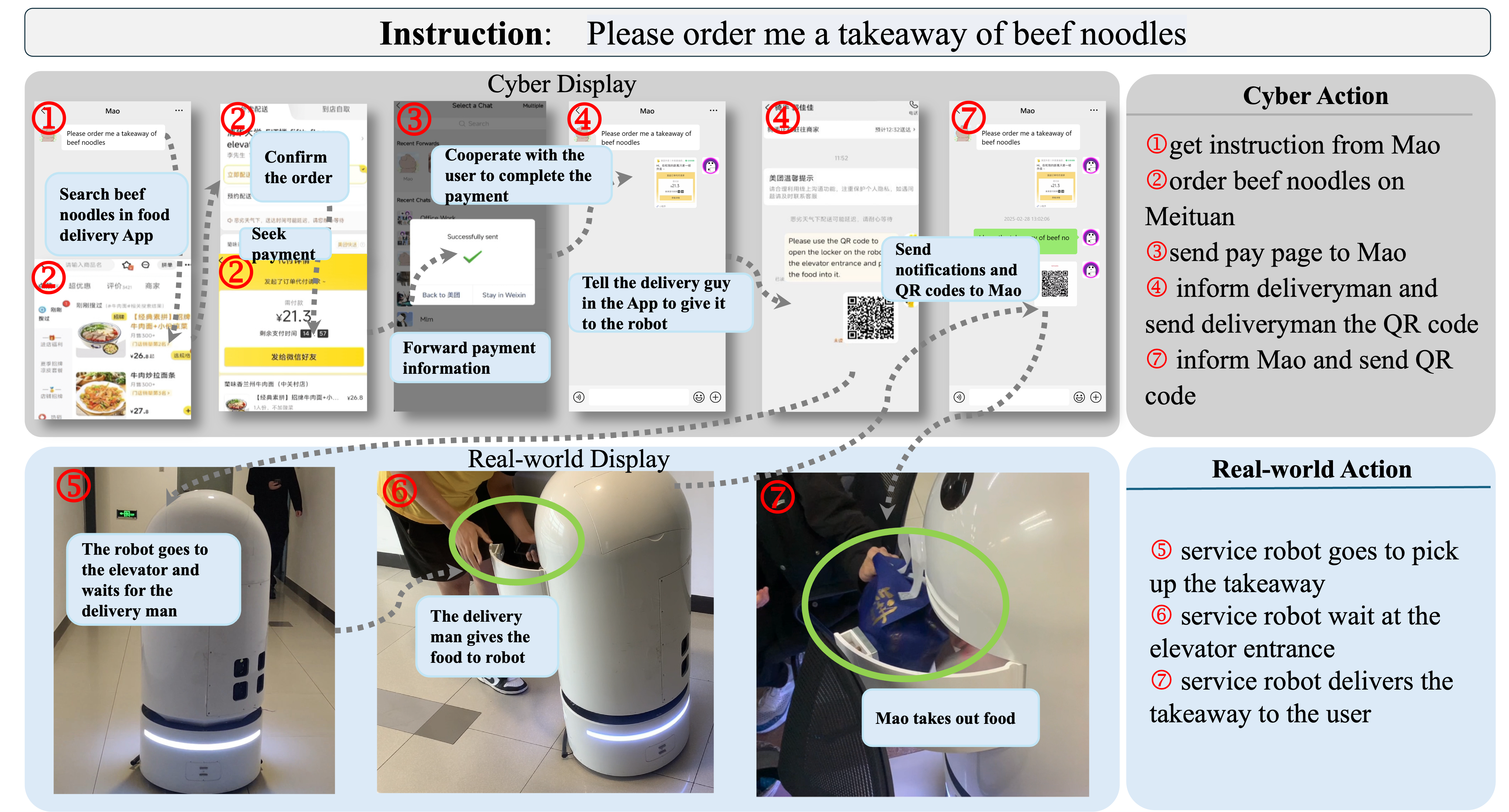}
    \vspace{-3mm}
    \caption{
    \textbf{ A hybrid digital–physical workflow.}  
Given the instruction “Please order me a takeaway of beef noodles,” InteractGen coordinates digital actions with real-world robot execution.  
\emph{Cyber Display:} InteractGen searches for the dish, confirms the order, forwards the payment page to the user, completes payment, informs the delivery person through the app, and sends status updates and QR codes to the user.  
\emph{Real-world Display:} the mobile robot navigates to the pickup point, receives the takeaway from the delivery person, and returns to deliver it to the requester.  
The sequence illustrates how InteractGen seamlessly integrates online service operations with physical robot actions to accomplish multi-step tasks across digital and real environments.
    }
    \vspace{-3mm}
    \label{fig:deployment}
\end{figure*}

\subsection{Social Utility and Real-World Impact}

Beyond technical performance, \textit{InteractGen} demonstrates clear real-world utility. Unlike systems restricted to scripted benchmarks or controlled labs, it operates in dynamic multi-user service workflows. Its structured Thought-of-Action planning, reflective validation, and proactive perception enable robust execution of long-horizon tasks—such as printing, signing, and document delivery—and consistently outperform decentralized baselines in both simulation and deployment.

This robustness translated directly into practice. In a three-month open deployment in a shared academic office, \textit{InteractGen} reliably handled natural user instructions, coordinated heterogeneous robots, and managed shared resources with minimal intervention. Users reported reduced physical effort and cognitive load, smoother collaboration, and improved inclusivity, as the system’s neutral request style mitigated interpersonal friction and encouraged mutual assistance.

Figure~\ref{fig:deployment} illustrates a representative example. We integrate \texttt{DBNet} and \texttt{ConvNeXtViT-document} for OCR, \texttt{GroundingDINO} for icon grounding, and \texttt{Qwen-VL-Plus} for GUI captioning to enable multimodal perception and GUI-level operation via simulated tapping. This allows workflows such as one-shot food ordering: the system navigates the interface, coordinates robotic delivery, and only prompts users for payment. The deployment also included wheeled and quadruped robots (Fig.~\ref{fig:deployment_2}), requiring minimal adaptation, demonstrating the generality of the overall architecture.

Together, these results highlight both the technical and social significance of \textit{InteractGen}. Technically, it provides a scalable and interpretable multi-agent reasoning framework that supports reliable, adaptive execution in dynamic environments. Socially, it acts as a coordination mediator that distributes tasks more fairly, reduces collaboration burden, and improves accessibility for users with limited mobility or time. This dual impact positions \textit{InteractGen} not merely as an automation system, but as a step toward sustainable, human-centered teamwork suitable for workplaces, assistive settings, and other socially embedded domains.

\section{Conclusion}

This work re-examines the role of foundation models in robotics through the lens of real-world service workflows. Rather than asking whether a single Vision–Language Model or Vision–Language–Action policy can serve as a universal robot brain, we showed that the core limitation is architectural: monolithic models are fundamentally misaligned with the distributed, evolving, and dependency-heavy nature of human-populated environments. Building on this perspective, we introduced InteractGen, an LLM-powered multi-agent framework that treats foundation models as regulated components within a structured control system rather than as end-to-end embodied policies. By decomposing robot intelligence into specialized agents for perception, planning, assignment, validation, reflection, and human delegation, InteractGen demonstrates that reliable autonomy emerges from coordinated roles, shared memory, and explicit feasibility checks. The framework delivers consistent gains across simulation and real-world deployment, including a three-month open-use study in a shared office, where it executed long-horizon tasks with high reliability, adapted fluidly to environmental changes, and collaborated with humans in natural and socially appropriate ways. Even low-cost mobile robots exhibited markedly elevated capability when embedded in InteractGen’s human-aware multi-agent loop, handling tasks that far exceed the reach of standalone foundation models.

At the same time, our study has limitations that point to rich future directions. InteractGen is currently constrained by hand-engineered agent roles and does not explicitly demonstrate emergent abilities. Extending the framework to learn skills, role decompositions, and inter-agent protocols directly from data would further improve scalability and broaden domain coverage. Our experiments focus on domestic–office workflows with moderate team sizes; deploying similar architectures in safety-critical domains such as healthcare, eldercare, or public spaces will require tighter integration with formal safety guarantees, richer uncertainty modeling, and stronger robustness to adversarial or strategically behaving humans. Moreover, while our results highlight the social utility of treating humans as active agents, a deeper understanding of long-term trust, fairness, and workload distribution in human–robot–agent collectives remains an open challenge.

We view InteractGen not as a final answer, but as a compelling existence proof that foundation models, when embedded within carefully structured multi-agent control systems, can substantially advance service robotics toward intentional, human-centered autonomy with measurable real-world impact. More broadly, the framework illustrates how robots and LLM-driven agents can move beyond reactive execution toward collaborative problem solving with people—supporting clearer communication, more adaptive task sharing, and safer, more trustworthy human–robot interaction. We believe such systems mark a step toward assistive technologies that amplify human capability, reduce cognitive and physical burden, and enable more equitable access to real-world robotic assistance.

\bibliographystyle{IEEEtran}
\bibliography{reference}

\end{document}